\documentclass[journal,web,twoside]{IEEEtran}

\usepackage{color}

\ifCLASSINFOpdf
  \usepackage[pdftex]{graphicx}
\else
\fi
\usepackage[hoptionsi]{overpic}

\usepackage{amsmath}
\usepackage{bm}
\usepackage{amsfonts}
\usepackage{amssymb}

\usepackage{algorithm}
\usepackage{algorithmic}

\usepackage{makecell}

\ifCLASSOPTIONcompsoc
  \usepackage[caption=false,font=normalsize,labelfont=sf,textfont=sf]{subfig}
\else
  \usepackage[caption=false,font=footnotesize]{subfig}
\fi

\usepackage{multirow}
\usepackage{threeparttable}
\usepackage{adjustbox}

\begin{document}
%
\title{A Unified Framework for Generalized Low-Shot Medical Image Segmentation with Scarce Data}
%
%
%

\author{Hengji~Cui,
        Dong~Wei,
        Kai~Ma,
        Shi~Gu,
        and~Yefeng~Zheng,~\IEEEmembership{Senior Member,~IEEE}
\thanks{Manuscript received July 1, 2020; revised October 27, 2020; accepted December 14, 2020.
This work was primarily supported by the NSFC General Program 61876032, the Key-Area Research and Development Program of Guangdong Province, China (No. 2018B010111001), the National Key R\&D Program of China (2018YFC2000702), and the Scientific and Technical Innovation 2030-``New Generation Artificial Intelligence''  Project (No. 2020AAA0104100).
\emph{(H. Cui and D. Wei contributed equally to this work.)
(Corresponding author: S. Gu.)}}
\thanks{H. Cui and S. Gu are with the School of Computer Science and Engineering, University of Electronic Science and Technology of China (email: chj@std.uestc.edu.cn, gus@uestc.edu.cn). H. Cui contributed to this work as an intern at Tencent Jarvis Lab.}
\thanks{D. Wei, K. Ma, and Y. Zheng are with Tencent Jarvis Lab (email: donwei@tencent.com, kylekma@tencent.com, yefengzheng@tencent.com).}}%

%
%

\markboth{Published in IEEE TRANSACTIONS ON MEDICAL IMAGING,~Vol.~40, No.~10, October~2021}%
{Cui \MakeLowercase{\textit{et al.}}: A Unified Framework for Low-Shot Medical Image Segmentation with Scarce Data}
%



\maketitle

\begin{abstract}
Medical image segmentation has achieved remarkable advancements using deep neural networks (DNNs).
However, DNNs often need big amounts of data and annotations for training, both of which can be difficult and costly to obtain.
In this work, we propose a unified framework for generalized low-shot (one- and few-shot) medical image segmentation based on distance metric learning (DML).
Unlike most existing methods which only deal with the lack of annotations while assuming abundance of data, our framework works with extreme scarcity of both, which is ideal for rare diseases.
Via DML, the framework learns a multimodal mixture representation for each category, and performs dense predictions based on cosine distances between the pixels' deep embeddings and the category representations.
The multimodal representations effectively utilize the inter-subject similarities and intraclass variations to overcome overfitting due to extremely limited data.
In addition, we propose adaptive mixing coefficients for the multimodal mixture distributions to adaptively emphasize the modes better suited to the current input.
The representations are implicitly embedded as weights of the fc layer, such that the cosine distances can be computed efficiently via forward propagation.
In our experiments on brain MRI and abdominal CT datasets, the proposed framework achieves superior performances for low-shot segmentation towards standard DNN-based (3D U-Net) and classical registration-based (ANTs) methods, \emph{e.g.}, achieving mean Dice coefficients of 81\%/69\% for brain tissue/abdominal multi-organ segmentation using a single training sample, as compared to 52\%/31\% and 72\%/35\% by the U-Net and ANTs, respectively.
\end{abstract}


%
\IEEEpeerreviewmaketitle

\section{Introduction}
%
%
%
%
\IEEEPARstart{S}{emantic} segmentation, one of the fundamental tasks in computer vision, has important applications in medical image analysis, such as risk assessment~\cite{klifa2010magnetic}, 
precise diagnosis~\cite{silveira2009comparison}, therapy planning~\cite{jackson2018deep}, 
and clinical research.
Benefiting from the rapid development of deep learning techniques, deep neural network (DNN) based methods have advanced the state of the art of medical image segmentation---by large margins over traditional methods---in recent years~\cite{litjens2017survey}.
To obtain satisfactory performance, large amounts of data with annotations are usually needed for training DNNs.
However, accurate annotation of medical images is expertise-demanding, labor-intensive, time-consuming, and error-prone, thus difficult and costly to obtain in large amount.
Moreover, in some circumstances, \emph{e.g.}, for rare conditions,
even the training data can be very limited.

To cope with the lack of training samples, low-shot (one- and few-shot) learning techniques~\cite{fei2006one} 
have been proposed in the natural image domain and achieved encouraging results in both classification~\cite{snell2017prototypical,vinyals2016matching} 
and segmentation tasks~\cite{wang2019panet,zhang2019canet}.
In the low-shot natural image domain, it is often assumed (and also thus practiced) that generalizable prior knowledge can be learned on a sufficiently large set of labeled samples of known \emph{base} classes, and then utilized to boost learning of previously unseen \emph{novel} classes given limited samples (\emph{i.e.}, the target task).
A few works on low-shot medical image \emph{classification} followed this problem setting of base versus novel classes.
For example, Paul \emph{et al.}~\cite{Paul2020fast} divided different chest diseases into disjoint base (with large training set) and novel (with a few training samples per class) classes, and proposed a few-shot learning framework based on autoencoder ensemble for disease identification from chest x-ray images.
For many \emph{segmentation} tasks of medical images, however, the problem setting of low-shot learning is usually different from the assumption above.
Taking one-shot segmentation of brain structures in MRI for example~\cite{zhao2019dataaug}:
the concepts of base and novel classes are infeasible here, as all the classes to segment are already presented in the given one-shot example;
or in other words, there is no unseen novel class.
Consequently, most low-shot segmentation methods proposed for natural images cannot be directly applied to medical images.

Alternatively, most recent works on DNN-based low-shot medical image segmentation~\cite{zhao2019dataaug,chaitanya2019semi,wang2020LTNet} relied on image synthesis.
These methods utilized unlabeled data of the target task in semi-supervised setting, assuming abundance of data.
For example, both Zhao \emph{et al.}~\cite{zhao2019dataaug} and Chaitanya \emph{et al.}~\cite{chaitanya2019semi} proposed to learn and utilize spatial and appearance transformations between the labeled and unlabeled samples to synthesize realistic samples for training the segmentation networks.
Although requiring only one or few annotations, these semi-supervised methods~\cite{zhao2019dataaug,chaitanya2019semi,wang2020LTNet} require a considerable amount of unlabeled data to work effectively~\cite{tajbakhsh2020embracing}, thus inapplicable when
the samples are rare, too.
The scenario in which both the annotations and data are scarce can be practically relevant, \emph{e.g.}, when developing algorithms for a rare disease, and when the emergence of a novel disease necessitating rapid algorithm development.
However, it is more challenging due to the shortage in both data and annotations.
Indeed, such scenario is rarely touched in existing literature on DNN-based approaches.
In contrast, registration-based segmentation~\cite{avants2011reproducible}---a classical approach to few-shot segmentation of medical images---can achieve decent performance using only one or few labeled samples (called ``atlas''), especially for brains.
Nonetheless, this approach can become less effective when large individual variations in appearance and/or structures
are present and is usually time-consuming, too.


In this work,
we propose a new DNN-based framework for generalized low-shot medical image segmentation using extremely scarce training samples and annotations---a full analogy to humans who can learn reasonably well in such situation, which is able to handle large individual variations and accomplish segmentation in seconds.
Unlike most existing DNN-based methods for low-shot medical image segmentation, which relied on image synthesis, the framework employs the distance metric learning (DML)~\cite{karlinsky2019repmet} to learn a generalized multimodal representation (comprising one to several prototypes) for each category.
The DML of multimodal representation can effectively utilize not only the inter-subject similarities, but also the intraclass variations, for robustness against overfitting in case of scarce data.
In the context of medical image data and for a specific imaging modality, the inter-subject similarity refers to the similar appearance of the same tissue, organ, \emph{etc.}, across different individuals, whereas the intraclass variations refer to the intrinsic heterogeneity of a specific organ/tissue.
Notably, we propose to adaptively determine the mixing coefficients for the modes of a category based on a self-attention mechanism~\cite{hu2018squeeze}, rather than treating all modes equally or relying on only one of them for prediction~\cite{karlinsky2019repmet}.
Further, we implement the multimodal representations using the cosine normalization technique~\cite{luo2018cosine}, by implicitly embedding the representations as weights of the fully connected (fc) layer and computing the cosine similarity by forward propagation.
Compared to explicit representation learning~\cite{snell2017prototypical,karlinsky2019repmet}, the implicit embedding not only simplifies the network structure but also promotes computational efficiency.
Accordingly, we name our framework the Multimodal Representation Embedding Network (MRE-Net).
Last but not least, the MRE-Net handles both one- and few-shot learning with a unified network structure.

Besides the multimodal representation embedding (MRE),
we further employ three strategies targeted at dealing with the unique characteristics of medical image data.
First, the online hard example mining (OHEM)~\cite{shrivastava2016training} technique
is adopted in our dense prediction task to deal with the class imbalance problem commonly encountered in medical images.
Second, patch-wise spatial information~\cite{liao2019evaluate} is incorporated for utilization of structural similarities among individuals.
Third, the channel-wise attention~\cite{hu2018squeeze} and atrous spatial
pyramid pooling (ASPP)~\cite{chen2017rethinking} are employed to account for the diversity of the segmentation targets in medical image analysis.

In summary, our contributions are three folds:
\begin{itemize}
  \item First, we propose the MRE-Net, a unified framework for generalized low-shot (one- and few-shot) medical image segmentation in case of scarcity of both annotations and samples.
      The MRE-Net employs the DML for multimodal category representation learning to effectively utilize the inter-subject similarities and intraclass variations.
  \item Second, we propose adaptive mixing coefficients for the modes of a category, a key innovation in multimodal representation learning to ensure that the modes more suitable for the current input are given more emphasis.
  \item Third, we conduct thorough experiments on two distinct datasets (brain and abdominal multi-organ~\cite{BTCV,xu2016evaluation} segmentation) to demonstrate the superiority of the MRE-Net to both classical and modern methods in a variety of low-shot settings.
      The behaviour of the MRE-Net is studied thoroughly, too, for additional insights.
\end{itemize}

\section{Related Work}

\subsubsection{Semantic Segmentation}
Semantic segmentation refers to the process of assigning each pixel in an image a class label.
In terms of medical image segmentation, these labels may include organs (\emph{e.g.}, heart, lung, and liver), tissues (\emph{e.g.}, brain structures), bones, just to name a few.
Fully convolutional networks (FCNs)~\cite{long2015fully} 
are the current state of the art for semantic segmentation, which replaced the fc layers at the end of the network with convolutional layers to output pixel-level dense prediction.
Arguably, U-Net and its variants~\cite{ronneberger2015u,luna20183d} 
(also belonging to FCNs) may be the most widely used structure for medical image segmentation, which features symmetric downsampling and upsampling paths (the encoder and decoder) with skip connections in between.
Although being the \emph{de facto} benchmark for medical image segmentation, U-Nets are data-hungry just like most DNNs.
In this work, we use a modified 3D U-Net~\cite{luna20183d} as the backbone feature extractor, as well as a baseline in the experiments.
As will be shown in Section~\ref{sec:experiments}, low-shot performance of the U-Net can be substantially improved after enhanced by the proposed MRE and trained in the way of DML.

\subsubsection{DML and Low-shot Segmentation of Natural Images}
DML has been widely used in computer vision applications (see~\cite{kulis2012metric} for a survey). 
Recently, this concept has been revitalized in the realm of deep learning~\cite{kaya2019deep} and become one of the central topics in few-shot learning~\cite{snell2017prototypical,wang2019panet,zhang2019canet,karlinsky2019repmet}.
In DML, a learnable embedding function projects the input into an embedding space, where the projected samples of the same class should be closer to each other than those of different classes.
To learn the embedding function, a distance metric (\emph{e.g.}, squared Euclidean~\cite{snell2017prototypical,karlinsky2019repmet} and cosine distances~\cite{wang2019panet,qi2018low}) is employed to measure the intra- and inter-class distances, and loss functions are defined accordingly (\emph{e.g.}, the triplet loss~\cite{weinberger2009distance}) to encourage small intra-class and large inter-class distances.
Commonly in low-shot segmentation,
masked average pooling is performed~\cite{wang2019panet,zhang2019canet} to obtain category representations, and segmentation labels of individual pixels are obtained by dense comparison against the representations.

As introduced earlier,
it is prohibitive to directly apply low-shot segmentation methods for natural images to medical images in practice.
However, we find the idea of DML-based representation learning powerful, and apply it to low-shot medical image segmentation.
Notably, our work is most related to the RepMet~\cite{karlinsky2019repmet} and weight imprinting~\cite{qi2018low}: the former proposed multimodal representations for low-shot detection, and the latter showed that cosine similarity based DML could be equivalently implemented by implicitly embedding category representations as weights of the fc layer.
In RepMet, each category was represented by a Gaussian mixture model with multiple modes, whose centers were considered as the representative vectors for the category;
however, only one mode was used for a category at the time of classification, and the explicit representations in RepMet unnecessarily complicated the network structure.
As to \cite{qi2018low}, its unimodal weight embedding did not fully exploit the intraclass variations.
In this work, we innovatively implement multimodal representation learning by extending the weight embedding technique from single mode to multiple modes, and adaptively emphasize the most informative modes with a self-attention mechanism~\cite{hu2018squeeze}.

\subsubsection{Low-shot Medical Image Segmentation}
Most existing works on DNN-based low-shot medical image segmentation relied on image synthesis.
Zhao \emph{et al.}~\cite{zhao2019dataaug} proposed to learn both spatial and appearance transformations from an annotated atlas to a pool of unlabeled images, and synthesized labeled data by combining randomly sampled transformations.
The synthesized data were then used as augmented data to train a segmentation DNN.
Similarly, Chaitanya \emph{et al.}~\cite{chaitanya2019semi} employed two conditional generative adversarial nets (GANs)~\cite{mirza2014conditional} to synthesize augmented data using spatial and additive intensity transformations.
GAN~\cite{goodfellow2014generative} was also employed in \cite{mondal2018few} for few-shot segmentation of 3D multi-sequence MR images, using labeled, unlabeled, and fake images for semi-supervised training.
Wang \emph{et al.}~\cite{wang2020LTNet} introduced the forward-backward consistency to the DNN-based registration framework~\cite{balakrishnan2019voxelmorph}, which enabled a variety of extra, cycle-consistency-based supervision signals to make the training process stable, while also boosting the performance of one-shot segmentation.
All the works mentioned above belong to the genre of semi-supervised methods, which require a considerable amount of unlabeled data to work effectively, hence inapplicable to the scenario considered in this work.

One exception to the above-reviewed works is the more recent `channel squeeze \& spatial excitation' (sSE) guided method~\cite{roy2020squeeze}, which followed the problem setting of base and novel classes in the natural image domain~\cite{shaban2017one}.
Concretely, Roy \emph{et al.}~\cite{roy2020squeeze} used three abdominal organs as the base classes on which the proposed model was trained, and evaluated the model on the fourth organ (the unseen novel class).
A two-arm architecture~\cite{shaban2017one} consisting of a conditioner arm and a segmenter arm was adopted, with the former processing the annotated input, generating task-specific representations, and then passing these representations to the latter for task-specific segmentation guidance.
An sSE module was proposed to facilitate the information transfer between the two arms.
Encouraging results were reported on contrast-enhanced abdominal CT images.
Its principle difference with our work is the problem setting.
In \cite{roy2020squeeze}, a considerable amount of data (65 scans and corresponding annotations) of the base classes were used for training.
In contrast, we consider the case in which the data are scarce (one to several examples), \emph{e.g.}, for a rare condition.
Another difference is that \cite{roy2020squeeze} was intended for binary segmentation of a novel class, whereas our MRE-Net is for multiclass segmentation of all the ``base'' classes.

Registration-based segmentation~\cite{avants2011reproducible} is a classical approach to low-shot medical image segmentation, and can achieve decent performance using only one or few atlases, especially for brains.
Nonetheless, this approach can become less effective when large individual variations in appearance and/or structures
are present, and is usually time-consuming.
We propose the MRE-Net to address these two issues simultaneously.

\subsubsection{Data Augmentation}
Online data augmentation has become a standard technique for alleviating overfitting by DNNs, since first extensively used by Krizhevsky \emph{et al.} \cite{krizhevsky2012imagenet}.
It applies random transformations to the images
during training to artificially increase the diversity of the training data.
For medical images, commonly conducted transformations include mirroring, brightness/constrast jittering, rotation, and elastic deformation.
In this work, we empirically explore the impact of data augmentation in the low-shot setting.

\section{Methods}
\subsection{Conceptual Overview: Segmentation via DML-Based Dense Prediction}
\label{sec:method:overview}
\begin{figure}[t]
\centering
\includegraphics[width=\columnwidth,trim=130 160 110 150,clip]{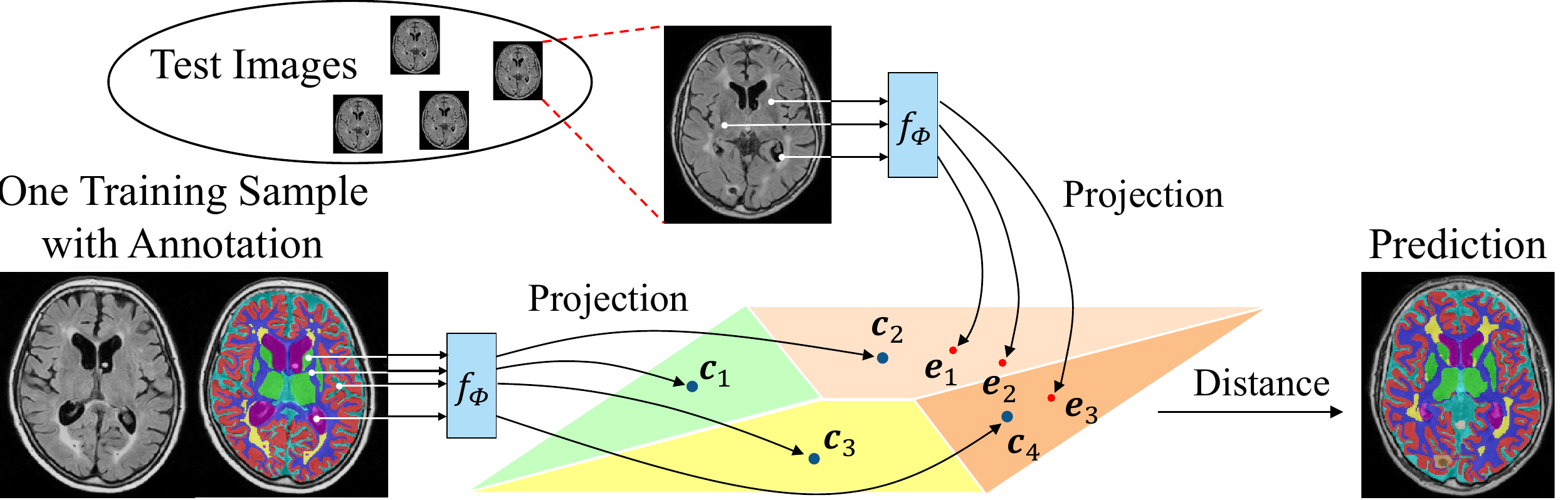}
\caption{Conceptual overview of our low-shot segmentation framework.
Pixels of test images are projected in the embedding space, where dense predictions are made according to the distances between pixels' embeddings $\bm{e}_i$ and category prototypes $\bm{c}_k$.
}\label{fig:concept}
\end{figure}

Given an annotated training set $\mathcal{X}^\mathrm{tn}=\{(x^{\mathrm{tn},(l)},a^{(l)})\}_{l=1}^{L}$, where $x^{\mathrm{tn},(l)}$ is the $l$\textsuperscript{th} training sample with the annotation $a^{(l)}$, and $L$ is the total number of annotated training samples,
we aim to infer the segmentation of any test image $x^\mathrm{ts}$ based on $\mathcal{X}^\mathrm{tn}$.
As conceptually illustrated in Fig.~\ref{fig:concept}, our framework first employs an embedding function $f_\Phi$ with learnable parameters $\Phi$ to project every pixel of the test image into the embedding space: $\bm{e}_i=f_\Phi(x^\mathrm{ts}_i|x^\mathrm{ts})$, where $i$ is a pixel index, and $\bm{e}_i\in\mathbb{R}^{N_{\bm{e}}\times1}$ is the resulting embedding vector.
Within the embedding space, a representative prototype $\bm{c}_k$
is learned from $\mathcal{X}^\mathrm{tn}$ for each segmentation category $k\in\{1,\ldots,K\}$, where $K$ is the total number of categories (including background).
With the notion of DML, the segmentation label for the pixel $x_i^\mathrm{ts}$ can be inferred non-parametrically by finding the nearest neighbor of $\bm{e}_i$ among all $\bm{c}_k$'s (although a more sophisticated mechanism than nearest neighbor is employed in our practical implementation).
In this work, we mainly focus on $L\leq3$, which is generalized low-shot segmentation.
Below, we describe our practical implementation of the DML and feature embedding process in detail.

\begin{figure*}[t]
\centering
\includegraphics[width=\textwidth,trim=0 102 0 25,clip]{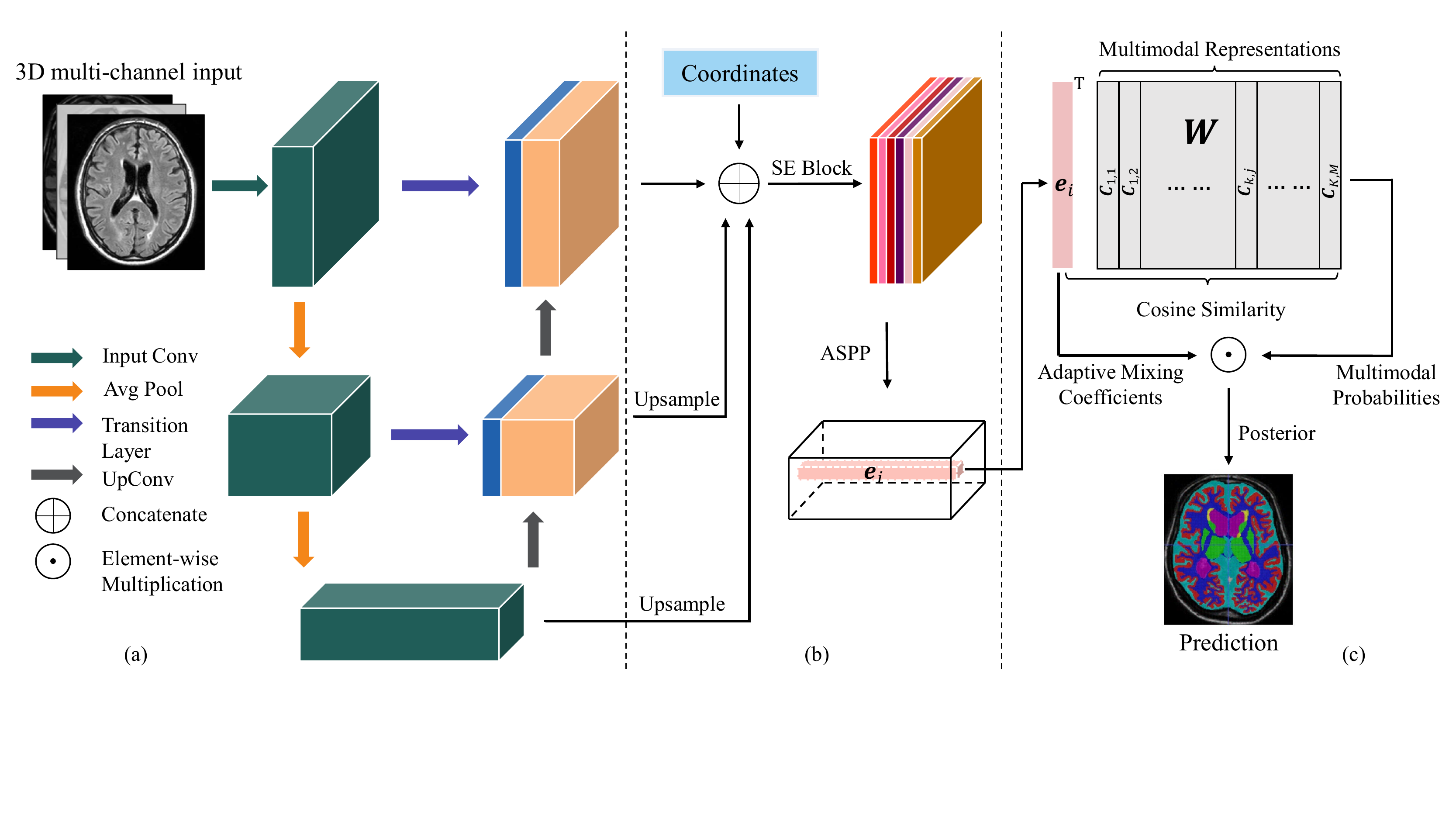}
\caption{Overview of the proposed framework:
(a) backbone network,
(b) dense feature embedding, and
(c) cosine distance-based dense prediction.
Note that 2D images are used for illustration purpose here, but the real data used in this work are 3D.
}\label{fig:struct}
\end{figure*}

\subsection{Backbone Network}
U-Net~\cite{ronneberger2015u} and its variants are widely used for medical image segmentation.
Luna and Park~\cite{luna20183d} used a modified 3D U-Net and won the first place in the Grand Challenge on MR Brain Segmentation at MICCAI 2018 (MRBrainS18)\footnote{https://mrbrains18.isi.uu.nl/}.
Most notably, to mitigate the demanding need for a large amount of 3D training data, a structure shallower than the original U-Net was designed;
in addition, the skip connection in the original U-Net was enhanced with a transition convolution, which rebalanced the low-level and high-level features by adjusting the number of channels in the former before concatenation.
For its outstanding performance on the same dataset as used in this work, we adapt this structure as our backbone feature extractor, and a strong baseline in the experiments.

Structure of the backbone network is illustrated in Fig.~\ref{fig:struct}(a). 
The input to the network are cropped 3D patches.
Multi-sequence data are treated as multichannel input.
The encoder comprises three blocks, consisting of three, four, and five convolutions, respectively.
The first convolutions in these blocks increase the channels of the feature maps to 64, 128, and 256, respectively.
The two average pooling layers between the encoder blocks reduce the size of the feature maps with the strides of [2, 2, 2] and [1, 2, 2] ([depth, height, width]), respectively.
Skip connections in the original U-Net~\cite{ronneberger2015u} are modified to transition layers, by adding channel-reducing convolutions (to 1/4 of the original) to the skip connections.
Each of the two blocks in the decoder includes two convolutions, the first of which reduces the channels of the feature maps to 128 and 64, respectively.
The transposed convolutions (UpConv) preceding the decoder blocks upsize the feature maps with the strides of [1, 2, 2] and [2, 2, 2], respectively.
The upsized feature maps are concatenated with the output of the transition layers, and processed by the decoder blocks.
All convolutions are followed by a rectified linear unit (ReLU) and batch normalization (BN).
In addition, all convolutions use the kernel size of 3$\times$3$\times$3, except for the transposed convolutions which have kernel sizes equal to those of their strides.
The final output is a 64-channel feature map of the same size as the input.

\subsection{Dense Feature Embedding}
From the perspective of FCNs~\cite{long2015fully}, segmentation is actually pixel-wise dense classification.
In this sense, we have obtained a 64-dimensional feature vector for each pixel using the U-Net backbone.
For normal DNN-based segmentation with abundant training samples and annotations, these features often produce decent results.
In case of one-shot segmentation, however, they are not sufficiently descriptive of each pixel, given that the training data not only are limited in amount but also lack diversity in composition (\emph{i.e.}, all examples of a category are from a single/few subjects).
To offset this disadvantage, we propose to excavate more information from the backbone.
Concretely, we concatenate the feature maps (upsampled if needed) on the right side of the U-Net (Fig.~\ref{fig:struct}(b)) to collect features at different levels.
Different from the semantic segmentation in natural images, spatial information has strong implications in medical images as many tissues and organs are location-specific.
Hence, we further concatenate Cartesian coordinates as additional channels to the feature maps (see the ``Coordinates'' rectangle at the top of Fig.~\ref{fig:struct}(b)).
Specifically, we calculate a location map for each input image.
The location map has the same size as the feature map, and three channels corresponding to the normalized coordinates in the $x$, $y$, and $z$ axes, respectively.
The maximal and minimal values in each axis are normalized to 1 and $-$1, respectively.

Considering the diversity in the segmentation targets in medical image analysis, we adopt two countermeasures to make our proposed MRE-Net more robust to the diversity.
First, we feed the concatenated feature maps through an SE attention block~\cite{hu2018squeeze}.
The SE block adaptively recalibrates channel-wise feature responses by explicitly modelling interdependencies between channels.
Thus, the channels most informative about the specific segmentation task are emphasized.
Second, we employ an atrous spatial pyramid pooling (ASPP) module~\cite{chen2017rethinking} with five spatial scales for obtaining multi-scale features.
The five scales are: one 1$\times$1$\times$1 convolution, three 3$\times$3$\times$3 convolutions with the atrous rates of 6, 12, and 18, respectively, and the image-level features.
The outputs of the ASPP module are concatenated and fused by another 1$\times$1$\times$1 convolution to produce the final pixel-wise embedding
$\bm{e}_i$.
Together, the concatenation of the (upsampled) multilevel feature maps, SE attention, and ASPP are deemed a holistic enhancement to the original U-Net feature embedding.
We abbreviate the enhanced feature embedding as the attentional multi-scale (AMS) embedding, whose effect will be studied later using an ablation experiment.

\subsection{DML of Multimodal Representation via Weight Embedding}
To deal with potentially serious overfitting due to the extreme scarcity of training data, we employ DML~\cite{karlinsky2019repmet} to learn category-discriminative representations for distance-based pixel-wise dense prediction,
which makes good use of the inter-subject similarity.
In addition, to fully exploit the intraclass variations, we employ a multimodal representation---instead of a single prototype---for each category, which is expected to further improve model generalization against overfitting.
The explicit category representations employed in the previous work~\cite{karlinsky2019repmet} caused a cumbersome network structure;
in addition, only a single mode was used for classification at a time,
thus the intraclass variations captured by the multimodal representations were not fully utilized.
In contrast, we present a multimodal representation learning framework with self-adaptive mixing coefficients to simultaneously simplify the network design, enhance computational efficiency, and fully exploit the intraclass variations.

\subsubsection{DML via Weight Embedding}\label{sec:method:DML:weight_emb}
For simplicity, we first introduce the case of unimodal category representation for DML, before describing our multimodal framework.
Following the notation introduced in Section~\ref{sec:method:overview}, the distance between a pixel $x_i$'s embedding vector $\bm{e}_i$ and a category prototype $\bm{c}_k$ can be computed by a function $d(\bm{e}_i,\bm{c}_k)$.
Then, the probability of $x_i$ belonging to the $k$\textsuperscript{th} category is proportional to the exponential of the negative distance:
$
p_ {k} (x_i)\propto \exp \left[-d(\bm{e}_i,\bm{c}_{k}) \right].
$
The discriminative class posterior $\mathbb{P}$ can be computed with a softmax operation over all candidate categories:
\begin{equation}
\mathbb{P} ( s_i = k ) = \frac{p_{k}(x_i)}{\sum_{k=1}^K p_{k}(x_i)} \triangleq
\frac{\exp\left[-d(\bm{e}_i,\bm{c}_{k})\right]}
{\sum_{k=1}^K\exp\left[-d(\bm{e}_i,\bm{c}_{k})\right]},
\end{equation}
where $s_i$ is the segmentation label of $x_i$.
Consider the cosine distance, \emph{i.e.}, $
d(\bm{e}_i,\bm{c}_{k})=1-\cos(\theta)
=1-\big(\frac{\bm{e}_i}{\|\bm{e}_i\|}\big)^\mathrm{T} \big(\frac{\bm{c}_k}{\|\bm{c}_k\|}\big)
=1-\hat{\bm{e}}_i^\mathrm{T}\hat{\bm{c}}_k
$, where $\theta$ is the angle between $\bm{e}_i$ and $\bm{c}_{k}$, $\|\cdot\|$ is the L2 norm, and $\hat{\bm{e}}_i$ and $\hat{\bm{c}}_k$ are the L2-normalized pixel's embedding vector and category prototype, respectively;
it can be easily derived that $\frac{\exp(-1+\hat{\bm{e}}_i^\mathrm{T}\hat{\bm{c}}_k)}
{\sum_k\exp(-1+\hat{\bm{e}}_i^\mathrm{T}\hat{\bm{c}}_k)}=
\frac{\exp(\hat{\bm{e}}_i^\mathrm{T}\hat{\bm{c}}_k)}
{\sum_k\exp(\hat{\bm{e}}_i^\mathrm{T}\hat{\bm{c}}_k)}$.
The operation $\hat{\bm{e}}_i^\mathrm{T}\hat{\bm{c}}_k$ can be readily implemented using an fc layer with all biases set to zero, where $\bm{c}_k$ is the weight vector connected to the $k$\textsuperscript{th} output node, and each node represents a category.
In this way, the cosine distance-based metric learning can be implicitly implemented with the commonly used fc layer, where the category prototypes are embedded as the weights of the fc layer: $\bm{W}=[\bm{c}_1, \ldots, \bm{c}_k]\in\mathbb{R}^{N_{\bm{e}}\times K}$, instead of using explicit representations.
From a different perspective, Luo \emph{et al.}~\cite{luo2018cosine} discussed the benefits of the cosine similarity in terms of preactivation computation, including more flexibility and lower test error, and described the computing process as ``cosine normalization'';
we refer interested readers to~\cite{luo2018cosine} for more details.
Closely related to our work, Qi \emph{et al.}~\cite{qi2018low} used the cosine normalization for unimodal representation learning;
in the next section we will describe how we apply the technique to multimodal representation learning and enhance it with self-adaptive mode-mixing coefficients.

Following~\cite{qi2018low}, we add a learnable scaling factor $\xi$ to adjust the exact values of the exponential function, thus we have:
\begin{equation}
\mathbb{P} ( s_i = k ) =
{\exp\left(\xi\hat{\bm{e}}_i^\mathrm{T}\hat{\bm{c}}_k\right)}
\Big/
{{\sum}_{k=1}^K\exp\left(\xi\hat{\bm{e}}_i^\mathrm{T}\hat{\bm{c}}_k\right)}.
\end{equation}

\subsubsection{Multimodal Weight Embedding with Adaptive Mixing Coefficients}\label{sec:method:weights}
We now consider the case of multimodal representation learning via the cosine normalization technique (Fig.~\ref{fig:struct}(c)), and more importantly, how to determine the relative contribution of each mode to the posterior $\mathbb{P}$.
Specifically, we represent each category by a mixture distribution model, where the centers of the modes are the category prototypes (\emph{i.e.}, a category is represented by a set of prototypes, instead of a single one).
Let us denote the prototypes of the $k$\textsuperscript{th} category by
$R_k=\{\bm{c}_{k,j}\}_{j=1}^M$, where $\bm{c}_{k,j}$ represents the center of the $j$\textsuperscript{th} mode and $M$ is the number of modes.
Same as in the unimodal case, each prototype gives an individual probability
$
p_{k,j} (x_i)\propto \exp \left[-d(\bm{e}_i,\bm{c}_{k,j}) \right],
$
and the multimodal mixture distribution is defined as
$
p_k(x_i)={\sum}_{j=1}^M \alpha_{k, j}p_{k, j}(x_i),
$
where $\alpha_{k,j}$ is the mixing coefficient for the $j$\textsuperscript{th} mode of the $k$\textsuperscript{th} category, and $\sum_{j=1}^{M}\alpha_{k,j}=1$.
Then, the posterior $\mathbb{P}$ can now be computed using the mixture models:
\begin{equation}
\begin{aligned}
\mathbb{P} ( s_i = k ) &= \frac{p_{k}(x_i)}{\sum_{k=1}^K p_{k}(x_i)}
= \frac{{\sum}_{j=1}^M \alpha_{k, j}p_{k, j}(x_i)}
{ {\sum}_{k=1}^K {\sum}_{j=1}^M \alpha_{k, j}p_{k, j}(x_i) }\\
&\triangleq \frac{\sum_{j=1}^M \alpha_{k,j} \exp\left(\xi\hat{\bm{e}}_i^\mathrm{T}\hat{\bm{c}}_{k,j}\right)}
{\sum_{k=1}^K\sum_{j=1}^M \alpha_{k,j} \exp\left(\xi\hat{\bm{e}}_i^\mathrm{T}\hat{\bm{c}}_{k,j}\right)}.
\end{aligned}
\end{equation}
Conveniently, the multimodal DML can still be implemented using an fc layer, whose biases and weight matrix are set to zero and $\bm{W}=[\bm{c}_{1,1}, \bm{c}_{1,2}, \ldots, \bm{c}_{k,j}, \ldots, \bm{c}_{K,M}]\in\mathbb{R}^{N_{\bm{e}}\times KM}$, respectively, where each category is jointly represented by $M$ output nodes and $\bm{c}_{k,j}$ is the weight vector connected to the $j$\textsuperscript{th} node of the $k$\textsuperscript{th} category.

The mixing coefficient $\alpha_{k,j}$ controls the relative contribution of each mode, thus is an important design consideration.
Karlinsky \emph{et al.}~\cite{karlinsky2019repmet} set $\alpha_{k,j}=1$ if $j=\mathop{\arg\max}_j p_{k,j}$ and $\alpha_{k,j}=0$ otherwise, \emph{i.e.}, setting $\mathbb{P}$ to the upper bound on the actual class posterior.
In this way, $\mathbb{P}$ was solely determined by the mode which was closest to the pixel's embedding vector for each category, thus the intraclass variations were not fully utilized.
One way to make use of all the modes is to set $\alpha_{k,j}=1/M$, \emph{i.e.}, treating each mode of a category equally.
However, the equal-weight setting implied an underlying assumption that $\bm{e}_i$ was equally similar to $\bm{c}_{k,j}$ for all $j$, which may not always hold.
In this work, we propose self-adaptable mixing coefficients for each category's multimodal distribution.
It is implemented as a modified ``squeeze-and-excitation'' (SE) self-attention mechanism~\cite{hu2018squeeze}, which allows end-to-end training and is simple and straightforward.
First, an fc layer takes $\bm{e}_i$ (of dimension $N_{\bm{e}}\times1$) as input and outputs a vector of dimension $512\times1$ (the ``squeeze''), where $N_{\bm{e}}\geq512$, followed by ReLu.
Then, another fc layer projects the squeezed and ReLu'ed vector to the dimension $KM\times1$ (the ``excitation''), followed by a sigmoid activation, producing a vector $\bm{\beta}=\big[\beta_{k,j} | k=1, \ldots, K, \mathrm{and}\ j=1, \ldots, M\big]$ (we denote the parameter of these two fc layers by $\theta_\mathrm{mix}$).
Lastly, the mixing coefficients are given by a per-category softmax operation:
\begin{equation}\label{eq:weights}
  \alpha_{k,j}={\exp(\beta_{k,j})}\Big/{{\sum}_{j=1}^{M}\exp(\beta_{k,j})}.
\end{equation}
As we can see, this process adaptively calculates the mixing coefficients for the modes depending on the actual input $\bm{e}_i$,
thus selectively emphasizing as well as neglecting different modes as needed.
It is worth noting that setting $M=1$ would degrade the framework to be unimodal.
Hence, the unimodal and multimodal settings are unified within a single framework.



\subsection{Loss Computation}
We use the per-category normalized cross-entropy loss as the dense prediction loss:
\begin{equation}\label{eq:loss}
  \mathcal{L}_\mathrm{CE} = -{\sum}_k\delta_k(a_i) \log\big[\mathbb{P}(s_i=k)\big]\big/N_k,
\end{equation}
where $a_i\in\{1, \ldots, K\}$ is the annotation label for pixel $i$, $\delta_k(a_i)$ is the Dirac delta function, which equals 1 if $k = a_i$ and 0 otherwise, and $N_k$ is the number of pixels of the $k$\textsuperscript{th} category that are used for loss computation in a mini-batch.
The category representations are implicitly embedded as weights of the fc layer, and updated by descending along the stochastic gradient of $\mathcal{L}_\mathrm{CE}$.
Hence, the framework learns the backbone network parameters, the embedding space, and the representations simultaneously in an end-to-end manner by optimizing $\mathcal{L}_\mathrm{CE}$.


Class imbalance is a common problem for medical images, which may result in under-sampling and unsatisfactory performance for categories with considerably fewer examples.
OHEM~\cite{shrivastava2016training} has shown outstanding ability in dealing with foreground/background imbalance in object detection tasks, thus we adapt it to work pixel-wise for our low-shot segmentation task.
Specifically, all the categories for segmentation are empirically grouped into a majority group and a minority group, based on the number of pixels in each category.
Then, in each mini-batch, we keep all the pixels from the minority group for the loss computation, and rank those from the other according to their loss values.
We keep only the most difficult majority-group pixels (identified by greatest losses) and discard the rest.
In practice, the number of the majority-group pixels to keep is set to an integer ($N_\mathrm{Lg}$) multiple of the total number of minority-group pixels.

\subsection{Training and Testing Procedures}
Training procedures of the proposed MRE-Net is listed in Algorithm \ref{alg:training}.
Here, the feature embedding function $f_\Phi$ includes both Fig.~\ref{fig:struct}(a) (the backbone network) and Fig.~\ref{fig:struct}(b) (the SE block and ASPP module).
As we can see, all the learnable parameters (\emph{i.e.}, $\Phi$, $\bm{W}$, and $\theta_\mathrm{mix}$) of the MRE-Net are updated together by training the network end-to-end.
For testing, we only need to forward the test image through the MRE-Net (following the arrows in Fig. \ref{fig:struct}), and the semantic label for a pixel can be determined by
\begin{equation}\label{eq:test}
  s_i = \operatorname{argmax}_k\mathbb{P}(s_i=k).
\end{equation}

\begin{algorithm}[!t]
  \caption{Training procedure of the proposed MRE-Net.}\label{alg:training}
  \begin{algorithmic}[1]
  \REQUIRE Annotated training set $\mathcal{X}^\mathrm{tn}=\{(x^{\mathrm{tn},(l)},a^{(l)})\}_{l=1}^{L}$, where $L\leq3$, and learning rate hyperparameter $\eta$
  \ENSURE Learned network parameters $\Omega=\{\Phi, \bm{W}, \theta_\mathrm{mix}\}$
  \STATE {Randomly initialize $\Omega$}
  \FOR {number of training iterations}
    \STATE Sample mini-batch of images (patches) from $\mathcal{X}^\mathrm{tn}$
    \STATE Evaluate $\mathcal{L}_\mathrm{CE}$ in Equation (\ref{eq:loss}) using the mini-batch
    \STATE Update parameters with gradient descent:
    \STATE $\Omega\leftarrow\Omega-\eta\nabla_\Omega\mathcal{L}_\mathrm{CE}$
  \ENDFOR
  \end{algorithmic}
\end{algorithm}

\section{Experiments}\label{sec:experiments}
\subsection{Datasets and Preprocessing}
We evaluate the proposed MRE-Net on two public datasets: the MRBrainS18 dataset and the Beyond the Cranial Vault (BTCV) Abdomen~\cite{BTCV,xu2016evaluation} dataset.

\subsubsection{MRBrainS18}
The task of the challenge is to segment various brain structures on multi-sequence brain MR images with and without (large) pathologies.
The publicly available data (\emph{i.e.}, the official training data) of MRBrainS18 comprises seven subjects, with fully annotated multi-sequence MRI scans acquired on a 3T scanner at the UMC Utrecht (the Netherlands), including T1-weighted (repetition time (TR): 7.9 ms and echo time (TE): 4.5 ms), T1-weighted inversion recovery (TR: 4416 ms, TE: 15 ms, and inversion time (TI): 400 ms), and T2-FLAIR (TR: 11000 ms, TE: 125 ms, and TI: 2800 ms) sequences.
Officially, the scans were bias field corrected using the N4ITK algorithm~\cite{tustison2010n4itk}, and the T1-weighted and T1-weighted inversion recovery images were aligned with the T2-FLAIR image for each subject.
As illustrated in Fig. \ref{fig:struct}(a), multiple sequences of a subject are concatenated as channels of the 3D input.
The manual annotations contain 10 labels excluding the background, and officially the evaluation is carried out on eight of them: cortical gray matter (GM), basal ganglia (BG), white matter (WM), white matter hyperintensities (WMH), cerebrospinal fluid (CSF), ventricles (VT), cerebellum (CB), and brain stem (BS).
The MRI volumes are of size $240\times240\times48$ voxels, with the voxel size of $0.958\times0.958\times3.0$ mm\textsuperscript{3}.
For computational efficiency, we crop the central region of size $216 \times216\times 48$ voxels for each scan, which is large enough to contain the whole brain.
The z-score standardization is conducted on each MRI volume to make the pixel intensities have a zero mean and unit standard deviation.
No skull stripping is performed.

\subsubsection{BTCV}
The dataset comprises abdominal CT scans acquired at the Vanderbilt University Medical Center
from metastatic liver cancer or post-operative ventral hernia patients.
The scans were captured during portal venous contrast phase with variable volume sizes
($512\times512\times85$--$512\times512\times198$ voxels) and field of views (approx. $280\times280\times280$ mm\textsuperscript{3}--$500\times500\times650$ mm\textsuperscript{3}).
The in-plane resolution varies from $0.54\times0.54$ mm\textsuperscript{2} to $0.98\times0.98$ mm\textsuperscript{2}, while the slice thickness ranges from 2.5 to 5.0 mm.
A total of 13 abdominal organs were manually annotated;
following~\cite{gibson2018automatic}, we report segmentation performances on a subset of eight organs: spleen, right kidney (r.kid.), left kidney (l.kid.), gallbladder, esophagus, liver, stomach, and pancreas.
For preprocessing, we crop the CT volumes to focus on the segmentation targets and reduce computational cost, according to the coordinates given in~\cite{gibson2018automatic};
the cropping coordinates for three scans (out of the 30 of which the annotations are publicly available) turn out problematic, reducing the total number of usable scans to 27.
After that, we resize the cropped volumes to $324\times396\times72$ voxels, and  truncate the Hounsfield units of all scans outside the range $[-210, 290]$ to ignore irrelevant image content.
Lastly, the image intensities are linearly scaled to $[0, 1]$.

A subset of seven scans is randomly selected for the low-shot experiments, mainly for two reasons.
First, this setting is consistent with that for the MRBrainS18 dataset, allowing direct comparison between results on both datasets in a common sense;
second, the remaining 20 scans will be used by a semi-supervised method as unlabeled auxiliary training data in a comparative experiment (to be described in Section \ref{sec:exp:DataAug}.)

\subsection{Experimental Settings}\label{sec:exp:setting}
We compare the performance of our proposed MRE-Net to that of the U-Net~\cite{luna20183d} (which ranked the first place in the MRBrainS18 challenge) in one-shot setting.
Concretely, the model is trained with one sample and tested with the other six, yielding a mean performance averaged over the six test samples for each segmentation category or the cross-category mean metric.
However, noting that the choice of the single labeled sample is crucial and can lead to a potential bias in the results, we repeat the experiment seven times by using each of the seven samples for training in turn, and report the mean performance and standard deviation computed over the seven runs.
We report the performance both with and without online data augmentation.
Also, we compare with the classical registration-based one-shot approach using the registration tool provided in the widely used Advanced Normalization Tools (ANTs)~\cite{avants2011reproducible}, which is one of the state-of-the-art methods for classical intensity-based registration.
As to the few-shot settings, we experiment with two- and three-shot learning for our MRE-Net, to gain insights into how it performs in such setting with increasing amount of training data.
Concretely, we randomly select pairs/triplets of samples for training and test on the remaining five/four samples.
Again, the experiments are repeated seven times and the mean results averaged over the seven runs are reported.

In addition, we evaluate the leave-one-out performance of the U-Net~\cite{luna20183d} on each dataset (\emph{i.e.}, each of the seven samples is used for testing in turn while the other six being used for training), which is served as the upper bound of the low-shot settings.
The Dice coefficient and 95\textsuperscript{th} percentile of the Hausdorff distance (HD95) are used as the evaluation metrics.
Statistical analysis is conducted to compare the mean performance of our MRE-Net with those of other approaches at the significance level of 5\% (with Bonferroni correction when applicable). 
For reference, we also present results reported in related, non-low-shot-learning works~\cite{luna20183d,gibson2018automatic} which also involved the two datasets used in this work, along with our experimental results.
However, due to the different amounts/compositions/sources of training and test data, backbone networks, and/or testing schemes, direct comparisons are precluded.


\begin{table}[t]
\caption{Details about training configurations.}\label{tab:implementation}
\centering
\setlength{\tabcolsep}{1.9mm}
\begin{tabular}{cccccc}
\Xhline{1.2pt}
& \multicolumn{2}{c}{MRBrainS18} & & \multicolumn{2}{c}{BTCV} \\ \cline{2-3}\cline{5-6}
& Ours & U-Net~\cite{luna20183d} & & Ours & U-Net~\cite{luna20183d}  \\ \Xhline{0.9pt}
$M$ & 3 & - & & 3 & - \\
$N_{\bm{e}}$ & 2048 & - & & 2048 & - \\
$N_\mathrm{Lg}$ & 6 & - & & 4 & - \\
Optimizer & \multicolumn{2}{c}{Adam~\cite{kingma2014adam}} & & \multicolumn{2}{c}{Adam~\cite{kingma2014adam}} \\
Learning rate (initial) & \multicolumn{2}{c}{0.001} & & \multicolumn{2}{c}{0.001}  \\
\multicolumn{2}{l}{Learning rate (adjustment)} & \multicolumn{4}{c}{Multiply by 0.1 every 1,500 iterations} \\
Input patch size & \multicolumn{2}{c}{$36\times36\times12$} & & \multicolumn{2}{c}{$36\times36\times12$} \\
Mini-batch size     & 24  & 64 &  & 24   &64  \\
Training iterations & 3,000 & 20,000 & & 3,000 & 20,000     \\
GPU device    & \multicolumn{5}{c}{NVIDIA TITAN Xp (12GB memory)}  \\
{No. of GPUs} & 4$\times$ & 1$\times$ & & 4$\times$ & 1$\times$  \\
\Xhline{1.2pt}
\end{tabular}
\end{table}

\subsection{Implementation}
The PyTorch framework is used for experiments.
We train the networks from scratch and until convergence.
The input are 3D patches randomly sampled from the given training samples.
When data augmentation is involved,
random left-right mirroring (50\% probability; MRBrainS18 only), and random brightness and contrast jittering are implemented.
For OHEM, the GM, WM, and VT are considered as the majority group for the MRBrainS18 dataset as they jointly occupy more than 85\% of brain tissues according to our statistics,
whereas gallbladder and esophagus are considered as the minority group for the BTCV dataset considering their small sizes.
Although for both datasets only a subset of all the annotation labels is used for evaluation, we still maintain representations for non-evaluation categories for better performance on target categories.
More training details are charted in Table~\ref{tab:implementation}.


\begin{table*}[!t]
\caption{Results of the ablation study on design components in one-shot setting. Values are mean (std.).}\label{tab:ablation}
\centering
\setlength{\tabcolsep}{3.55mm}
\begin{tabular}{ccccccccccc}
\Xhline{1.2pt}
&\multirow{2}{*}{DML}& Cartesian & AMS & \multirow{2}{*}{OHEM} && \multicolumn{2}{c}{MRBrainS18} && \multicolumn{2}{c}{BTCV} \\
\cline{7-8}\cline{10-11}
&& coordinate & embedding &&& Dice (\%) & HD95 (mm) && Dice (\%) & HD95 (mm) \\ 
\hline
(a) & - & - & - & - && 14.86 (19.11) & 59.30 (18.21) && 25.91 (5.53) & 58.29 (16.08) \\
(b) & \checkmark & - & - & - && 65.38 (2.27) & 14.47 (2.08) && 53.36 (2.75)	& 35.47 (9.26) \\
(c) & \checkmark & \checkmark & - & - && 69.41 (2.72) & 9.36 (1.34) && 59.08 (2.47) & 26.59 (8.70) \\
(d) & \checkmark & \checkmark & \checkmark & - && 74.47 (1.65) & 6.95 (1.77) && 67.86 (1.90) & 21.18 (8.99) \\
(e) & \checkmark & \checkmark & \checkmark & \checkmark && \textbf{78.39} (1.07) & \textbf{6.30} (1.73) && \textbf{69.13} (1.43) & \textbf{19.02} (8.63) \\
\hline
(f) & - & \checkmark & \checkmark & \checkmark && 38.17 (4.18) & 24.56 (3.54) && 31.23 (4.82) & 42.63 (12.69) \\
(g) & \checkmark & - & \checkmark & \checkmark && 70.67 (2.34) & 12.76 (2.18) && 56.96 (2.67) & 22.69 (10.22) \\
\Xhline{1.2pt}
\end{tabular}
\end{table*}

For testing, a straightforward sliding window approach is employed, in which the segmentation labels of consecutive patches of size $28\times28\times8$ voxels are predicted.
Notably, to avoid the unwanted boarder effects, the actual input to the trained network is an expanded patch of size $36\times36\times12$ voxels and only the prediction results for the central $28\times28\times8$ voxels are used.
Being aware that more sophisticated testing approaches are likely to produce better results (\emph{e.g.}, Luna and Park~\cite{luna20183d} predicted every voxel eight times in their testing scheme), we choose the simple approach to focus on demonstrating the effectiveness of our proposed network infrastructure in generalised low-shot medical image segmentation, instead of pursuing extreme performance by adopting complicated implementations.

For a fair comparison, we use the same loss function and testing scheme for the U-Net and the proposed MRE-Net.

\begin{table}[!t]
\caption{Impact of the embedding dimension $N_{
\bm{e}}$ (with $M=3$).
Values are mean (std.), where applicable.}\label{tab:dimension}
\centering
\setlength{\tabcolsep}{.85mm}
\begin{tabular}{ccccc}
\Xhline{1.2pt}
$N_{\bm{e}}$ & 512 & 1024 & 2048 & 4096 \\ 
\hline
Dice (\%)  & 75.77 (1.93)	& 77.56 (0.95)	& \textbf{78.39} (1.07)	& 76.99 (1.26) \\
HD95 (mm)  & 6.59 (1.98) & 6.62 (1.62) & \textbf{6.30} (1.73) & 6.47 (1.86) \\
GPU memory (GB) & $\sim$4$\times$5.90 & $\sim$4$\times$7.68 & $\sim$4$\times$10.18 & $\sim$4$\times$14.44 \\
\Xhline{1.2pt}
\end{tabular}
\end{table}

\begin{table}[!t]
\caption{Impact of the number of modes $M$ for each category (with $N_{\bm{e}}=2048$).
Values are mean (std.).}\label{tab:num_mode}
\centering
\setlength{\tabcolsep}{.7mm}
\begin{adjustbox}{width=1.\linewidth}
\begin{tabular}{cccccc}
\Xhline{1.2pt}
$M$ & 1 & 2 & 3 & 4 & 5 \\
\Xhline{0.7pt}
\multicolumn{6}{c}{One-shot}\\
\hline
Dice (\%)  & 74.91 (1.98) & \textbf{78.42} (1.13) & 78.39 (\textbf{1.07}) & 78.05 (1.30) & 77.68 (1.62) \\
HD95 (mm)  & 7.08 (1.81) & \textbf{6.21} (1.55) & 6.30 (1.73) & 6.54 (1.83) & 6.93 (\textbf{1.40}) \\
\Xhline{0.7pt}
\multicolumn{6}{c}{Three-shot}\\
\hline
Dice (\%)  & 79.21 (1.25) & 81.39 (1.43) & \textbf{81.98} (\textbf{1.11}) & 80.99 (1.39) & 81.23 (1.91) \\
HD95 (mm)  & 5.88 (1.83) & \textbf{5.25} (1.79) & 5.26 (\textbf{1.69}) & 5.44 (1.76) & 5.50 (1.75) \\
\Xhline{1.2pt}
\end{tabular}
\end{adjustbox}
\end{table}

\subsection{Important Design Options}
Using the MRBrainS18 dataset, we empirically study two important design options of the proposed MRE-Net: dimensionality of the feature embedding $N_{\bm{e}}$ and the number of modes $M$ for each category.
(i) We vary $N_{\bm{e}}$ in the set $\{512, 1024, 2048, 4096\}$, and chart the one-shot performance and GPU memory consumption during training in
Table~\ref{tab:dimension}.
As expected, the memory consumption increased markedly with the increase of $N_{\bm{e}}$.
Meanwhile, the performance improved noticeably from $N_{\bm{e}}=512$ and peaked at $2048$.
Combining both factors, we select $N_{\bm{e}}=2048$ for other experiments.
(ii) We vary $M$ from 1 to 5 in both one- and three-shot settings, and present the corresponding performance in Table~\ref{tab:num_mode} (the GPU memory consumption is relatively stable thus not included).
As we can see, the trends were similar for both settings: the performance first increased and then fluctuated as $M$ increased.
Taking into account both the means and standard deviations of both metrics in both settings, we select $M=3$ for experiments in this paper.

\begin{table}[!t]
\caption{Impact of Mixing Coefficients.
Values are mean (std.).}\label{tab:mixing_coefficients}
\centering
\setlength{\tabcolsep}{2.9mm}
\begin{tabular}{cccc}
\Xhline{1.2pt}
 & One-hot~\cite{karlinsky2019repmet} & Average & Adapt (proposed) \\ 
\hline
Dice (\%) & 77.04 (1.28) & 76.08 (1.38) & \textbf{78.39} (1.07) \\
HD95 (mm) & 7.59 (1.96) & 6.54 (1.78) & \textbf{6.30} (1.73) \\
\Xhline{1.2pt}
\end{tabular}
\end{table}

\subsection{Impact of Mixing Coefficients}
Using one-shot learning on the MRBrainS18 dataset, we compare the performance of different approaches to setting the mixing coefficients of the multimodal representations (described in Section~\ref{sec:method:weights}), \emph{i.e.}, setting that of the nearest prototype to 1 and others to 0 (``one-hot'')~\cite{karlinsky2019repmet}, using all equal coefficients (``average''), and adapting the coefficients according to the input (``adapt'').
The results are presented in Table~\ref{tab:mixing_coefficients}.
As we can see, the proposed adaptive approach achieves the best performance in terms of both the Dice and HD95 metrics.

\subsection{Ablation Study on Design Components}
Considering inherent characteristics of medical image data, we innovatively incorporate four targeted solutions in our framework:
(i) DML of multimodal representations to utilize inter-subject similarity and intra-class variations against overfitting,
(ii) incorporation of spatial information (the Cartesian coordinates) to utilize inter-subject structural similarity,
(iii) the attentional multi-scale (AMS) embedding for diverse medical images,
and (iv) pixel-level OHEM to overcome the class imbalance problem.
To verify the effectiveness of the proposed solutions, we conduct an ablation study
by incrementally adding them to a baseline network to gradually build up our framework.
The results are shown in Table~\ref{tab:ablation}, from which we notice similar trends on the MRBrainS18 and BTCV datasets.
Therefore, in the following we only describe the results on the former for conciseness.
As we can see, the introduction of the DML brought improvements over 300\% in both Dice and HD95 over the baseline.
On top of that, the Cartesian coordinates, AMS embedding, and OHEM brought further conspicuous incremental improvements in both metrics.
To summarize, all the proposed components contributed to the good performance of our framework, with the DML being most significant.

\begin{table*}[!t]
\caption{Results of the ablation study on distance metrics in one-shot setting.
Values are mean (std.).}\label{tab:ablation_distance}
\centering
\setlength{\tabcolsep}{2.35mm}
\begin{tabular}{ccccccccccc}
\Xhline{1.2pt}
& \multicolumn{4}{c}{MRBrainS18} && \multicolumn{4}{c}{BTCV} \\
\cline{2-5}\cline{7-10}
Distance & Dice (\%) & HD95 (mm) & GPU memory & Training time && Dice (\%) & HD95 (mm) & GPU memory & Training time \\ 
\Xhline{.6pt}
Cosine & 78.39 (1.07) & 6.30 (1.73) & $\sim$4$\times$10.18 GB & $\sim$150 min && 69.13 (1.43) & 19.02 (8.63) & $\sim$4$\times$9.81 GB & $\sim$140 min \\
Euclidean & 77.57 (1.01) & 6.52 (1.91) & $\sim$8$\times$11.30 GB & $\sim$260 min && 67.68 (1.48) & 18.95 (9.17) & $\sim$8$\times$10.95 GB & $\sim$250 min \\
\Xhline{1.2pt}
\end{tabular}
\end{table*}

\begin{table*}[t]
\caption{Experimental results on the MRBrainS18 dataset (``Aug.'' stands for data augmentation; ``-\emph{n}" means $n$-shot learning; ``Mean excl.'' represents results excluding WMH).
Values are mean (std.).
Bold faces denote best results per column (grouped by one- and few-shot settings);
asterisks (*) denote statistically significant differences from the best results in ``Mean'' and ``Mean excl.'' columns.}\label{tab:MRBrainS18}
\setlength{\tabcolsep}{.65mm}
\centering
\begin{adjustbox}{width=1.\linewidth}
\begin{threeparttable}
\begin{tabular}{ccccccccccc}
\hline
\Xhline{1.2pt}
& GM & BG & WM & WMH & CSF & VT & CB & BS & Mean (std.) & Mean excl. (std.)\\ \Xhline{1pt}
\multicolumn{11}{c}{\itshape Dice Similarity Coefficient (\%)}\\ \hline
ANTs-1~\cite{avants2011reproducible} & 57.06 (3.18) & \textbf{77.27} (1.80) & 63.98 (4.16) & 26.36 (3.01) & 57.29 (2.69) & 85.13 (3.49) & 84.07 (2.79) & \textbf{76.24} (11.25) & *65.93 (1.39) & *71.58 (1.29) \\
U-Net-1~\cite{luna20183d}& 14.06 (25.56) & 6.44 (12.58) & 22.22 (23.88) & 0.18 (0.27) & 24.63 (20.72) &  18.77 (23.22) & 16.35 (20.56) & 16.21 (13.79) & *14.86 (19.11) & *16.96 (18.83) \\
Aug. + U-Net-1~\cite{luna20183d}& 58.84 (13.40) & 41.54 (9.46) & 56.56 (15.93) & 26.38 (7.22) & 63.13 (11.97) &  56.53 (15.33) & 55.50 (11.27) & 34.72 (13.43) & *49.15 (11.91) & *52.40 (12.28) \\
MRE-Net-1 (ours) & \textbf{80.84} (1.61) & 74.88 (2.71) & \textbf{83.70} (1.29) & 58.42 (2.03) 	& \textbf{79.50} (1.58) & \textbf{90.57} (4.43) & 87.57 (1.76) & 71.66 (6.22) & \textbf{78.39} (1.07) & \textbf{81.25} (1.01) \\
Aug. + MRE-Net-1 & 80.60 (1.83) 	& 75.16 (1.95) & 83.21 (1.30) & \textbf{58.82} (1.39) & 79.25 (1.63) & 90.32 (3.06) & \textbf{87.60} (2.13) & 71.61 (5.49) & 78.32 (0.93) & 81.11 (1.04) \\

\hline

MRE-Net-2 (ours) & 82.34 (1.03) 	&78.94 (1.39) 	&84.78 (0.97) 	&68.67(14.26) 	&76.77 (0.66) 	& \textbf{92.23} (3.77) & 88.61 (0.84) 	&72.06 (2.42) 	&80.55 (1.31) 	&82.25 (1.23) \\
Aug. + MRE-Net-2 & 82.46 (1.24) 	&78.82 (1.30) 	&84.92 (1.06) 	&68.38 (13.99) 	&76.67(0.63) 	&91.92 (3.72) 	&88.46 (0.75) 	&71.78 (2.72) 	&80.43 (1.78) 	&82.15 (1.61) \\
MRE-Net-3 (ours) &82.85 (1.00) 	&78.53 (1.20) 	& \textbf{85.21} (1.13) & 72.64 (4.43) & \textbf{81.56} (0.24) & 91.22 (0.75) & \textbf{90.51} (0.74) & \textbf{73.32} (3.01) & \textbf{81.98} (1.11) & \textbf{83.31} (1.15) \\
Aug. + MRE-Net-3 & 82.42 (0.88) 	&78.67 (1.32) 	&84.86 (1.01) 	&73.42 (4.17) 	&81.14 (0.22) 	&91.07 (0.79) 	&90.11 (0.75) 	&73.03 (2.52) 	&81.84 (1.28) 	&83.04 (1.33) \\
U-Net-6~\cite{luna20183d} & 81.26 (2.17) &78.49 (2.69) &81.98 (3.48) &74.10 (10.21) &79.20 (1.54) &86.80 (2.57) &88.53 (2.09) &70.39 (5.30) &80.09 (2.76) &80.95 (3.17) \\
Aug. + U-Net-6~\cite{luna20183d} & \textbf{84.20} (1.94) & \textbf{78.96} (2.60) & 84.80 (2.74) 	& \textbf{76.09} (10.25) & 78.73 (1.39) 	&89.18 (3.77) 	&88.95 (1.78) 	&68.48 (5.69) 	&81.17 (2.60) 	&81.90 (2.80) \\

\hline
Non-low-shot~\cite{luna20183d}\tnote{a} & 86.0 (-) & 83.4 (-) & 88.2 (-) & 65.2 (-) & 83.7 (-) & 93.1 (-) & 93.9 (-) & 90.5 (-) & - & - \\

\Xhline{1pt}

\multicolumn{11}{c}{\itshape 95\% Hausdorff Distance (mm)}\\ \hline
ANTs-1~\cite{avants2011reproducible} & 3.04 (0.75) & \textbf{4.39} (0.40) & 3.93 (0.98) & 15.88 (0.86) & 3.86 (0.51) & 6.78 (6.59) & \textbf{5.01} (1.15) & \textbf{11.62} (9.86) & 6.81 (1.09) & \textbf{5.52} (1.17) \\
U-Net-1~\cite{luna20183d}& 42.16 (28.74)& 75.39 (29.41) & 40.69 (22.27)& 78.74 (23.61)& 33.78 (27.65)& 80.36 (20.30)& 71.33 (17.21)& 51.93 (4.55)& *59.30 (18.21) & *56.52 (17.97) \\
Aug. + U-Net-1~\cite{luna20183d} & 25.37 (20.17) &28.11 (18.52) &15.64 (16.96) &52.90 (19.16) &22.35 (11.33) &40.77 (13.36) &31.26 (9.09) &29.48 (10.34) & *30.74 (12.71) & *27.57 (11.96) \\
MRE-Net-1 (ours) & 2.28 (0.12) &5.36 (1.58) &2.74 (0.19) & \textbf{10.33} (0.76) & \textbf{2.64} (0.22) & \textbf{6.31} (1.23) &7.56 (3.75) &13.16 (9.93) & \textbf{6.30} (1.73) &5.72 (1.76) \\
Aug. + MRE-Net-1 & \textbf{2.10} (0.15) &5.32 (1.46) & \textbf{1.51} (0.21) &10.51 (1.51) &2.73 (0.18) &6.50 (1.45) &7.76 (3.78) &13.23 (8.80) &6.34 (1.87) &5.74 (1.79) \\

\hline

MRE-Net-2 (ours) & 1.75 (0.13) &4.36 (0.32) &2.57 (0.20) &10.02 (2.50) &2.43 (0.14) &4.78 (0.53) &6.93 (2.45) &12.04 (2.01) &5.61 (1.06) &4.98 (1.24) \\
Aug. + MRE-Net-2 & 1.69 (0.09) &4.26 (0.38) &2.83 (0.21) &9.77 (1.46) &2.54 (0.12) &4.67 (0.79) &6.64 (2.93) &12.26 (4.90) &5.58 (1.62) &4.99 (1.57) \\
MRE-Net-3 (ours) & 1.60 (0.11) & \textbf{4.03} (0.44) &2.49 (0.11) &9.87 (1.37) &2.31 (0.11) &3.26 (1.35) &6.17 (1.99) &12.32 (2.37) &5.26 (1.69) & 4.60 (1.66) \\
Aug. + MRE-Net-3 & \textbf{1.54} (0.09) &4.16 (0.30) &2.64 (0.15) & \textbf{9.61} (1.29) & \textbf{2.27} (0.14) & \textbf{3.00} (0.83) & \textbf{6.12} (1.96) &12.24 (2.00) & \textbf{5.20} (1.72) & \textbf{4.57} (1.75) \\
U-Net-6~\cite{luna20183d} & 1.86 (0.30) &4.42 (0.57) &2.91 (0.35) &17.25 (4.13) &2.95 (0.18) &8.24 (4.74) &7.16 (5.90) &15.31 (6.94) &7.51 (3.37) &6.12 (3.17) \\
Aug. + U-Net-6~\cite{luna20183d} &1.65 (0.10) &4.12 (0.44) & \textbf{2.37} (0.19) &14.31 (5.32) &2.55 (0.09) &7.64 (3.91) &7.08 (4.77) & \textbf{10.32} (6.06) &6.26 (2.60) &5.10 (2.80) \\

\hline
Non-low-shot~\cite{luna20183d}\tnote{a} & 1.27 (-) & 3.64 (-) & 2.12 (-) & 10.25 (-) & 2.22 (-) & 2.81 (-) & 3.74 (-) & 3.92 (-) & - & - \\

\Xhline{1.2pt}
\end{tabular}
\begin{tablenotes}
\item[a] As reported in \cite{luna20183d} (missing numbers are represented by `-').
    Note that different datasets and testing schemes preclude direction comparisons.
\end{tablenotes}
\end{threeparttable}
\end{adjustbox}
\end{table*}

To further validate the critical role the DML plays in the MRE-Net, we conduct an additional ablation experiment in which the DML implementation is removed
(\emph{i.e.}, producing class posteriors in the way of FCNs)
while the other advanced components listed in Table~\ref{tab:ablation} are kept.
The results are shown in row (f) of Table~\ref{tab:ablation}.
As we can see, the results were unusually low compared to those in rows (b)--(e), emphasizing the indispensable role of the DML.
In addition, we similarly conduct an exclusion experiment for the Cartesian coordinates (row (g)).
After removing the Cartesian coordinates, the performance dropped substantially compared to the full model (row (e)).
The results double-confirmed the benefit of utilizing inter-subject structural similarity by incorporating spatial information.

\begin{figure*}[!t]
\centering
\begin{minipage}[c]{\textwidth}\scriptsize
\includegraphics[width=\linewidth,trim=99 70 112 70,clip]{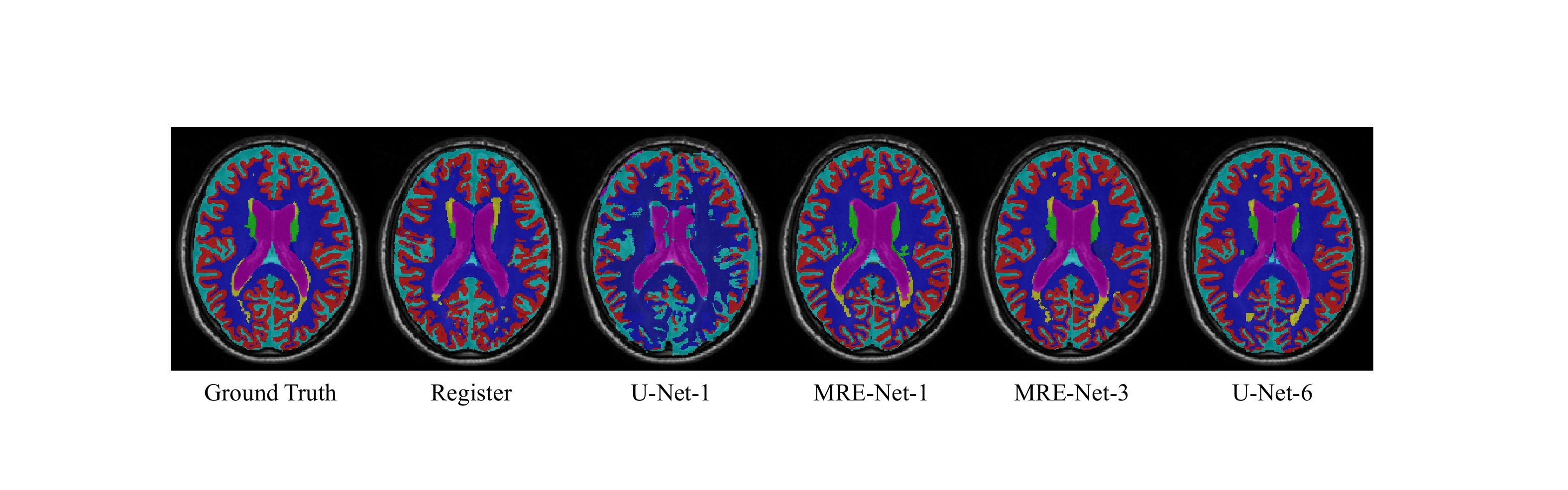}
\hspace*{7mm}Ground truth%
\hfill\hspace{7mm}ANTs~\cite{avants2011reproducible}\phantom{cd}%
\hfill Aug. + U-Net-1~\cite{luna20183d}%
\hfill MRE-Net-1 (ours)%
\hfill MRE-Net-3 (ours)%
\hfill Aug. + U-Net-6~\cite{luna20183d}\hspace*{3mm}
\end{minipage}
\caption{Segmentation results of different methods on the MRBrainS18 dataset.
Best viewed in color and zoomed-in.}\label{fig:vis}
\end{figure*}

\subsection{Ablation Study on Distance Metrics}
As introduced in Section \ref{sec:method:DML:weight_emb}, we use the cosine distance for the function $d(\bm{e}, \bm{c})$ to facilitate implementation of the DML with the fc layer.
In this section, we evaluate another distance---the squared Euclidean distance~\cite{karlinsky2019repmet}---in comparison with the cosine distance.
Specifically, it is defined as $d_\mathrm{Euc}(\bm{e}, \bm{c})=\big\|{\bm{e}}/{\|\bm{e}\|}-{\bm{c}}/{\|\bm{c}\|}\big\|^2=\|\hat{\bm{e}}-\hat{\bm{c}}\|^2$.
Implementation-wise, to use $d_\mathrm{Euc}$ we must represent the category prototypes with dedicated memory explicitly---which adds complexity to the framework, and can no longer compute the distances by forwarding through the fc layer as using the cosine distance.
Instead, $d_\mathrm{Euc}(\bm{e}, \bm{c})$ is densely computed on GPUs for exhaustive combinations of the pixel-wise embedding $\bm{e}$ and category prototype $\bm{c}$.
We compare the performance obtained using $d_\mathrm{Euc}$ on both the MRBrainS18 and BTCV datasets to that using the cosine distance;
besides, the computational resource consumption for training is compared, too.
The results are presented in Table \ref{tab:ablation_distance}.
On one hand, the performance of $d_\mathrm{Euc}$ is slightly lower than that of the cosine distance, but still comparable, \emph{e.g.}, the mean Dice coefficients on MRBrainS18 are 77.57\% (Euclidean) versus 78.39\% (cosine), while the corresponding HD95 are 6.52 mm (Euclidean) versus 6.30 mm (cosine).
On the other hand, using $d_\mathrm{Euc}$ consumes substantially more resources than using the cosine distance: it takes more than 2 times GPU memory and $\sim$1.7--1.8 times duration for training.
It is apparent that our choice of the distance metric (the cosine distance) and corresponding implementation (the cosine normalization technique~\cite{luo2018cosine}) are beneficial.

\begin{figure*}[t]
\centering
\begin{minipage}[c]{\textwidth}\scriptsize
\includegraphics[width=\linewidth,trim=45 325 204 130,clip]{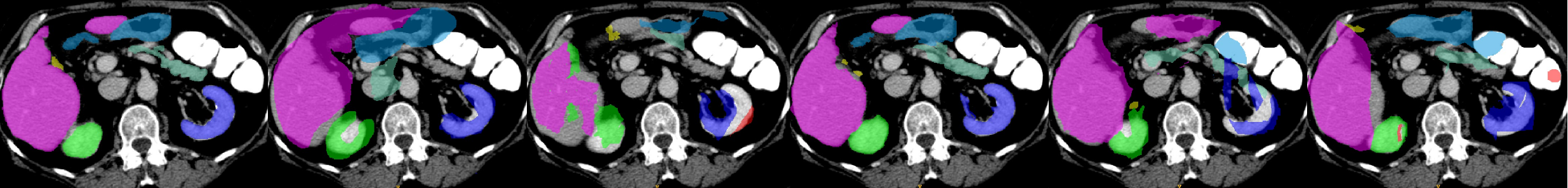}
\hspace*{8mm}Ground truth%
\hspace{17mm}ANTs-1~\cite{avants2011reproducible}%
\hspace{14mm}Aug. + U-Net-1~\cite{luna20183d}%
\hspace{12mm}MRE-Net-1 (ours)%
\hfill Aug. + U-Net-6~\cite{luna20183d}%
\hfill Aug. + U-Net-20~\cite{luna20183d}%
\hspace*{3mm}%
\end{minipage}
\caption{Segmentation results of different methods on the BTCV dataset.
Best viewed in color and zoomed-in.}\label{fig:vis_btcv}
\end{figure*}

\subsection{Evaluation Results on MRBrainS18}\label{sec:exp:MRBrainS18}
The results are presented in Table~\ref{tab:MRBrainS18}.
The WMH category is very challenging in that these are WM abnormalities without regular shape, size, or location.
Hence, we additionally present the mean evaluation metrics excluding WMH.
As we can see, the U-Net~\cite{luna20183d} trained with one sample (U-Net-1) without data augmentation performed poorly (mean Dice 14.86\% and HD95 59.30 mm).
With data augmentation, the performance of the U-Net improved markedly (mean Dice 49.15\% and HD95 30.74 mm), yet still unsatisfactory.
(Interestingly, the data augmentation in general improved the performance of the U-Net, yet did not cause much difference for the proposed MRE-Net.
Accordingly, we only describe the performance of the U-Net with the data augmentation, and that of the MRE-Net without it below.)
In contrast, the MRE-Net achieved quite reasonable results in the one-shot setting (mean Dice 78.39\% and HD95 6.30 mm), significantly outperforming those of U-Net-1.
When compared with U-Net-6, MRE-Net-1 achieved $\sim$97\% of the mean Dice of the upper bound (78.39$\pm$1.07\% vs. 81.17$\pm$2.60\%) and a comparable mean HD95 with that of the upper bound (6.30$\pm$1.73 mm vs. 6.26$\pm$2.60 mm).
Meanwhile, ANTs~\cite{avants2011reproducible} achieved decent one-shot performance: mean Dice 65.93\% and HD95 6.81 mm.
In comparison, the mean Dice and HD95 by MRE-Net-1 were higher by 12.46\% and smaller by 0.51 mm, respectively.

\begin{table*}[t]
\caption{Experimental results on the BTCV dataset (``Aug.'' stands for data augmentation, ``-\emph{n}" means $n$-shot learning).
Values are mean (std.).
Bold faces denote best results per column;
asterisks (*) denote statistically significant differences from the MRE-Net-1 in the ``Mean'' column.}\label{tab:BTCV}

\setlength{\tabcolsep}{.8mm}
\centering
\begin{threeparttable}
\begin{tabular}{cccccccccc}
\Xhline{1.2pt}
                        & Spleen      & R.kid.       & L.kid.      & Gallbladder     & Esophagus     & Liver       & Stomach       & Pancreas      & Mean (std.) \\ \Xhline{0.9pt}
\multicolumn{10}{c}{\itshape Dice Similarity Coefficient (\%)}\\ \hline
ANTs-1~\cite{avants2011reproducible} & 51.62 (2.97)&33.60 (4.63)&17.65 (5.57)&15.25 (4.25)&56.30 (3.71)&76.48 (1.55)&24.60 (1.61)&4.31 (0.64) & *34.98 (3.73) \\
Aug. + U-Net-1~\cite{luna20183d} & 31.48 (3.75)&	44.84 (4.11)&	34.03 (4.62)&	26.38 (2.98)&	17.24 (2.80)&	61.68 (1.41)&	11.56 (0.59)&	19.48 (0.82)&	*30.84 (5.22) \\
MRE-Net-1 (ours) & \textbf{78.08} (1.92) & \textbf{75.92} (1.70) & \textbf{77.43} (2.98) & \textbf{52.15} (1.76) & \textbf{56.83} (1.62)& \textbf{91.35} (0.81)& \textbf{59.65} (1.53) & \textbf{61.26} (2.21) & \textbf{69.13} (1.43) \\
\hline
Aug. + U-Net-6~\cite{luna20183d} & 63.49 (1.56)&	66.26 (1.89)&	66.29 (0.80)&	33.74 (2.32)&	30.99 (2.45)&	87.27 (0.88)&	33.19 (0.89)&	29.88 (1.02)&	*51.39 (2.98) \\
Aug. + U-Net-20~\cite{luna20183d}  & 64.81 (1.33)& 72.73 (1.77)&	69.26 (0.95)&	28.63 (1.12)&	25.75 (1.59)&	85.51 (1.04)&	45.95 (1.94)&	25.80 (0.43)&	*52.30 (2.57) \\
\hline
\multicolumn{1}{l}{Non-low-shot\tnote{a}} \\
\cline{1-1}
\multicolumn{1}{r}{Zhou \emph{et al.}~\cite{zhou2016three}} & 92 (-) & - & 91 (-) & 65 (-) & 43 (-) & 95 (-) & 60 (-) & 62 (-) & - \\
\multicolumn{1}{r}{DenseVNet~\cite{gibson2018automatic}} & 95 (-) & - & 93 (-) & 73 (-) & 71 (-) & 95 (-) & 87 (-) & 75 (-) & - \\
\Xhline{0.9pt}

\multicolumn{10}{c}{\itshape 95\% Hausdorff Distance (mm)}\\ \hline
ANTs-1~\cite{avants2011reproducible}   & 63.30 (10.57)&42.44 (13.59)&105.26 (29.81)&41.56 (16.76)&43.78 (10.11)&31.72 (7.70)&33.09 (5.89)&70.33 (7.43)& *53.93 (15.47) \\
Aug. + U-Net-1~\cite{luna20183d} & 47.14 (8.52)& 52.93 (10.71)& 51.77 (8.09)& 44.88 (7.98)& 40.39 (10.30)& 34.01 (5.27)& 32.91 (5.25)& 46.15 (6.17)& *43.77 (13.18) \\
MRE-Net-1 (ours)  & \textbf{22.93} (3.37) & \textbf{18.54} (3.20) & \textbf{19.47} (5.66) & \textbf{27.61} (7.23) & \textbf{15.87} (6.78) & \textbf{10.62} (2.78) & \textbf{12.98} (2.36) & \textbf{24.10} (5.16) & \textbf{19.02} (8.63) \\
\hline
Aug. + U-Net-6~\cite{luna20183d} & 43.05 (3.82)& 44.22 (5.00)& 47.33 (6.57)& 36.16 (9.39)& 35.06 (10.44)& 27.28 (7.17)& 28.45 (10.57)& 36.26 (6.69) & *37.23 (9.57) \\
Aug. + U-Net-20~\cite{luna20183d} & 42.57 (2.10)& 39.08 (4.31)& 46.65 (7.95)& 32.12 (5.12)& 34.87 (8.24)& 21.35 (7.48)& 20.74 (3.16)& 31.17 (6.19) & *33.57 (6.91) \\
\Xhline{1.2pt}
\end{tabular}
\begin{tablenotes}
\item[a] As reported in \cite{gibson2018automatic} (missing numbers are represented by `-');
    mean values for HD95 were not reported in \cite{gibson2018automatic}.
    Different datasets and testing schemes preclude direction comparisons.
\end{tablenotes}
\end{threeparttable}
\end{table*}

Regarding the few-shot setting, we can observe apparent improvements in both Dice and HD95 by increasing the number of training samples from one to three.
Notably, using two training samples, our MRE-Net already exceeded U-Net-6 in terms of mean HD95;
when using three training samples, the MRE-Net achieved a mean Dice higher than that of U-Net-6.
In summary, the MRE-Net achieved the best performance in both the one- and few-shot settings, demonstrating competence in low-shot brain MRI segmentation.
Example segmentation results of different methods are shown in Fig.~\ref{fig:vis}.

When compared with the non-low-shot approach proposed by Luna and Park~\cite{luna20183d}, we observe close performances of our MRE-Net and U-Net-6 in few-shot settings, although the non-low-shot approach yielded an overall superior performance.
In fact, all these three approaches employed the same 3D U-Net backbone, and the main differences were three-fold.
(i) Our MRE-Net incorporated countermeasures specific for low-shot learning---most notably the multimodal DML---on top of the backbone.
(ii) The non-low-shot approach employed a more sophisticated testing scheme for extreme performance in the challenge competition (and actually won the first place).
(iii) In addition, it was trained with all the seven subjects used in this work, and tested on 23 test subjects which are not publicly available.
With due caution, we try to explain the comparison as following.
First, the close performances validated our choice of the backbone.
Second, the overall superior performance of the non-low-shot approach to our MRE-Net might be partially attributed to the more complicated testing scheme, and the fact that more training data and different test data were used.
Third, we notice that for the WMH our evaluation results were actually better than those reported in \cite{luna20183d}, which might be caused by different distributions of the test data, urging caution in the interpretation, again.


\subsection{Evaluation Results on BTCV}
The proposed MRE-Net achieved competitive performance on the MRBrainS18 dataset.
To validate the capability of the MRE-Net to generalize for data of other modalities as well as other parts of the body, we also apply it to the BTCV dataset.
However, low-shot multi-organ segmentation in abdominal CT is more challenging, mainly for two reasons:
(i) the contrast of soft tissues in CT is not as good as that in MRI; and
(ii) the inter-subject variations in both appearance and structure---such as shape, size, and location---can be much larger in abdominal CT than in brain MRI.
From the preceding experiment on the MRBrainS18 dataset, we have learned that the data augmentation generally works for the U-Net but not for our MRE-Net.
Hence, we only experiment with the corresponding settings on the BTCV dataset.
The evaluation results are presented in Table~\ref{tab:BTCV}.
As we can see, our MRE-Net-1 achieved the best performances for all organ-wise and cross-organ evaluation metrics with large margins, substantially outperforming other competing methods including U-Net-6.
It is worth noting that neither ANTs nor the U-Net managed to produce reasonable results for this challenging task.
On the contrary, our framework was able to overcome the large inter-subject variations using only one sample for training.
To further evaluate the performance of MRE-Net-1 when compared to the U-Net trained with even more data, we train the U-Net with all the 20 scans not selected for the low-shot experiment and test it on the selected seven scans.
Results showed that U-Net-20 outperformed U-Net-6 in both mean Dice and HD95, as expected;
however, our MRE-Net-1 actually performed even better than U-Net-20.
We conjecture that the relatively small improvements of U-Net-20 over U-Net-6 might be partially attributed to the difficulty of the problem itself, and the possibility that these 20 scans did not present enough diversity for well training of the U-Net, which was not tailored for learning with limited data.
These results demonstrate superiority of our proposed MRE-Net in dealing with low-shot segmentation of data with higher extents of inter-subject variations.
Fig.~\ref{fig:vis_btcv} shows example segmentations produced by the different methods in Table~\ref{tab:BTCV}.

The performance of non-low-shot approaches to abdominal multi-organ segmentation reported in the literature~\cite{gibson2018automatic} was generally better than the one-shot performance of our MRE-Net, with the exceptions that the Dice values of MRE-Net-1 on esophagus, stomach, and pancreas were higher or on a par with those of \cite{zhou2016three}.
Again, different training and testing data were used: \cite{gibson2018automatic} used 90 CT scans (27 of which overlapped with the BTCV dataset used in this work) with a 9-fold cross-validation, and \cite{zhou2016three} used a totally different dataset of 240 CT scans with a 230:10 (train:test) split.
Therefore, the comparison results must be interpreted with due caution.
The superior evaluation results reported in \cite{gibson2018automatic,zhou2016three} to those obtained in this work were reasonable and expected, considering that many times more data were used for training.
Meanwhile, the superior performance of \cite{gibson2018automatic} to that of \cite{zhou2016three} might be partially attributed to the more advanced network structures of higher capacity employed in \cite{gibson2018automatic} than in \cite{zhou2016three}, although differences between the two datasets might also have played a role.

\begin{table}[!t]
\caption{Comparison of the time efficiency of different methods.}\label{tab:time}
\centering
\setlength{\tabcolsep}{1.05mm}
\begin{threeparttable}
\begin{tabular}{cccc}
\Xhline{1.2pt}
 & ANTs~\cite{avants2011reproducible} & Aug. + U-Net~\cite{luna20183d} & MRE-Net (ours) \\ 
\hline
Training (one-shot) & - & $\sim$120 min & $\sim$150 min \\
Inference (per sample) & $\sim$28 min\tnote{a} & $\sim$3s & $\sim$4s \\
\Xhline{1.2pt}
\end{tabular}
\begin{tablenotes}
\item[a] Using 64 threads on an Intel\textsuperscript{\textregistered} Xeon\textsuperscript{\textregistered} E5-2690 v4 @2.60GHz processor.
\end{tablenotes}
\end{threeparttable}
\end{table}

\subsection{Time Efficiency}
Table~\ref{tab:time} compares the time efficiency of the various methods in Table~\ref{tab:MRBrainS18}, including training time and average inference time, on the MRBrainS18 dataset.
Apparently, ANTs---despite yielding decent one-shot performance compared to the U-Net---was orders of magnitudes slower than the DNN-based approaches, kind of representative of the classical registration methods.
Meanwhile, despite 30-min longer training time, the proposed MRE-Net presents no practically perceivable difference in the inference time from the U-Net.

\subsection{Comparison with Existing Low-Shot Approaches}
As introduced earlier, existing approaches on few-shot medical image segmentation relied on the accessibility to auxiliary training data~\cite{zhao2019dataaug,roy2020squeeze}.
We are aware of the different usage scenarios intended, but also believe that it is useful to compare with some state-of-the-art (SOTA) methods among existing approaches.
In this part, we compare our proposed MRE-Net to the channel squeeze \& spatial excitation (sSE) guided method~\cite{roy2020squeeze} and data augmentation (DataAug) method~\cite{zhao2019dataaug}, with necessary adaption.
These two methods were both recently proposed, and representative of the one-shot medical image segmentation methods relying on (i) auxiliary data with annotations for base classes and (ii) unlabeled training data, respectively.
It is worth noting that although both methods have the potential for few-shot settings, the optimal way of extension is neither presented by the authors nor obvious.
Hence, the comparison is made in one-shot settings.

\subsubsection{Comparison with the sSE Method}
The sSE method~\cite{roy2020squeeze} was originally developed for the scenario in which a separate dataset with sufficient annotations of base classes is provided, to facilitate segmentation of a novel class versus the background (binary segmentation) given an example volume.
We adapt it for the one-shot scenario considered in this work, \emph{i.e.}, one-shot multiclass segmentation in which a single example with all target classes annotated is given but without any auxiliary data.
Concretely, we follow the training scheme of sSE to train the model until convergence, by sampling both the support and query sets from the only annotated volume.
For testing, the annotated volume is reused as the support volume to infer the segmentation of the query volumes, class by class.
We experiment with varying hyperparameters of sSE for optimal performance;
especially, we experiment with two different budgets of slice-wise annotations (the hyperparameter $k$): $k=10$ (the optimal tradeoff suggested in \cite{roy2020squeeze}) and $k=$ the total number of slices which contain a specific organ/tissue (the maximum value allowed, providing more guidance in theory);
the latter is enabled in our scenario as the support volume is fully annotated.
The same one-shot setting as described in Section \ref{sec:exp:setting} is used.

The results are charted in Table \ref{tab:sSE}.
We notice that increasing the number of slice-wise annotations from 10 to the maximum values allowed  did not result in substantial differences, which was consistent with the finding of Roy \emph{et al.} \cite{roy2020squeeze}.
However, our results also indicated that the sSE method with the straightforward adaption did not work for the target scenario considered in this work, yielding compromised performance.
We conjecture that the unsatisfactory performance was caused by severe overfitting, as sSE was originally designed to handle extreme scarcity of annotations rather than of data.


\begin{table}[!t]
\caption{Comparison with the sSE method~\cite{roy2020squeeze} in one-shot setting (``Aug.'' stands for data augmentation).
Values are mean (std.).
Bold faces denote best results per column, and asterisks (*) denote statistically significant differences from the MRE-Net.}\label{tab:sSE}
\centering
\setlength{\tabcolsep}{.5mm}
\begin{adjustbox}{width=1.\linewidth}
\begin{tabular}{cccccc}
\Xhline{1.2pt}
& \multicolumn{2}{c}{MRBrainS18} && \multicolumn{2}{c}{BTCV} \\
\cline{2-3}\cline{5-6}
 & Dice (\%) & HD95 (mm) && Dice (\%) & HD95 (mm) \\
\Xhline{.65pt}
MRE-Net (ours) & \textbf{78.39} (1.07) & \textbf{6.30} (1.73) & & \textbf{69.13} (1.43) & \textbf{19.02} (8.63) \\
Aug. + U-Net~\cite{luna20183d} & *49.15 (11.91) & *30.74 (12.71) & & *30.84 (5.22) & *43.77 (13.18) \\
Aug. + sSE ($k$ = 10) & *28.20 (2.26) & *31.58 (0.74) & & *23.04 (4.05) & *61.53 (13.14) \\
Aug. + sSE ($k$ = max.) & *22.66 (1.33) & *36.54 (0.90) & & *25.37 (1.75) & *63.87 (3.31) \\
\Xhline{1.2pt}
\end{tabular}
\end{adjustbox}
\end{table}

\begin{table*}[t]
\caption{Comparison with the semi-supervised DataAug method~\cite{zhao2019dataaug} in one-shot setting on the BTCV dataset (``Aug.'' stands for online data augmentation; ``-$n$'' means $n$ unlabeled scans were used).
Values are mean (std.).
Bold faces denote best results per column;
asterisks (*) denote statistically significant differences from the MRE-Net in the ``Mean'' column.}\label{tab:DataAug}

\setlength{\tabcolsep}{.8mm}
\centering
\begin{tabular}{crccccccccc}
\Xhline{1.2pt}
\phantom{abcdef} & & Spleen      & R.kid.       & L.kid.      & Gallbladder     & Esophagus     & Liver       & Stomach       & Pancreas      & Mean (std.) \\ \Xhline{0.9pt}
\multicolumn{11}{c}{\itshape Dice Similarity Coefficient (\%)}\\
\hline
\multicolumn{2}{c}{Aug. + U-Net~\cite{luna20183d}} & 32.32 (6.38) & 47.13 (18.40) & 32.40 (8.54) & 14.81 (4.14) & 19.43 (3.69) & 57.61 (11.90) & 54.25 (12.08) & 19.55 (5.88) & *34.69 (3.84) \\
\multicolumn{2}{c}{MRE-Net (ours)} & \textbf{78.76} (3.46) & \textbf{73.54} (4.42) & \textbf{78.36} (3.26) & \textbf{60.43} (9.77) & \textbf{55.44} (14.27) & \textbf{89.79} (3.08) & \textbf{62.43} (13.21) & \textbf{59.71} (15.94) & \textbf{69.81} (4.58) \\
\hline
\multicolumn{2}{l}{Semi-supervised~\cite{zhao2019dataaug}} \\
\cline{1-2}
& {SAS-aug-5} & 29.51 (3.44) & 8.10 (4.57) & 12.58 (4.91) & 6.85 (1.98) & 4.93 (3.08) & 60.78 (7.51) & 14.46 (9.68) & 18.28 (9.19) & *19.44 (3.15) \\
& {IDS-aug-5} & 36.26 (6.51) & 11.66 (6.33) & 15.58 (11.16) & 2.93 (4.49) & 13.59 (5.01) & 64.02 (9.29) & 23.49 (10.17) & 20.23 (8.23) & *23.47 (2.33) \\
\cline{2-11}
& {SAS-aug-10} & 38.76 (6.20) & 13.40 (3.93) & 13.30 (4.08) & 6.12 (3.72) & 9.51 (4.29) & 74.74 (3.12) & 24.08 (3.15) & 18.16 (8.87) & *24.76 (2.46) \\
& {IDS-aug-10} & 37.10 (8.11) & 12.85 (6.78) & 15.58 (11.94) & 3.46 (4.17) & 14.10 (5.65) & 64.61 (9.23) & 23.90 (9.83) & 21.84 (9.29) & *24.18 (2.61) \\
\cline{2-11}
& {SAS-aug-20} & 42.89 (8.13) & 16.51 (13.79) & 16.74 (8.27) & 6.68 (5.56) & 13.06 (2.98) & 79.53 (3.79) & 30.92 (3.88) & 22.34 (8.33) & *28.58 (3.74) \\
& {IDS-aug-20} & 38.85 (8.18) & 15.41 (6.86) & 16.97 (11.80) & 3.55 (2.86) & 17.63 (5.05) & 68.77 (7.59) & 28.98 (10.25) & 22.92 (10.20) & *26.64 (2.11) \\
\Xhline{0.9pt}

\multicolumn{11}{c}{\itshape 95\% Hausdorff Distance (mm)}\\
\hline
\multicolumn{2}{c}{Aug. + U-Net~\cite{luna20183d}} & 42.32 (15.49) & 51.76 (14.31) & 47.37 (24.62) & 46.31 (6.96) & 33.73 (11.83) & 34.15 (11.78) & 34.02 (9.10) & 39.26 (18.33) & *41.12 (7.43) \\
\multicolumn{2}{c}{MRE-Net (ours)} & \textbf{19.89} (9.73) & \textbf{16.57} (4.36) & \textbf{18.49} (5.63) & \textbf{21.02} (5.75) & \textbf{13.42} (6.11) & \textbf{12.31} (7.87) & \textbf{9.30} (5.05) & \textbf{15.37} (5.28) & \textbf{15.80} (3.46) \\
\hline
\multicolumn{2}{l}{Semi-supervised~\cite{zhao2019dataaug}} \\
\cline{1-2}
& {SAS-aug-5} & 43.08 (10.20) & 54.97 (7.92) & 57.66 (12.72) & 47.30 (13.55) & 37.58 (7.73) & 35.79 (7.21) & 39.78 (8.01) & 55.27 (13.45) & *46.43 (4.68) \\
& {IDS-aug-5} & 43.12 (7.36) & 45.40 (12.10) & 51.68 (5.49) & 47.19 (6.57) & 44.86 (7.08) & 31.65 (7.25) & 34.96 (12.72) & 43.20 (13.37) & *42.76 (3.74) \\
\cline{2-11}
& {SAS-aug-10} & 36.98 (9.83) & 49.37 (7.20) & 51.09 (12.16) & 42.15 (13.24) & 31.21 (7.94) & 30.34 (5.95) & 32.95 (7.75) & 47.98 (18.94) & *40.26 (3.67) \\
& {IDS-aug-10} & 42.40 (6.60) & 44.51 (12.52) & 49.07 (5.96) & 46.36 (9.74) & 44.03 (7.86) & 29.71 (6.56) & 33.04 (12.69) & 42.98 (13.82) & *41.51 (3.86) \\
\cline{2-11}
& {SAS-aug-20} & 34.86 (15.31) & 45.52 (13.15) & 47.00 (17.51) & 43.96 (6.13) & 29.15 (12.97) & 30.05 (14.22) & 27.32 (10.09) & 45.57 (24.33) & *37.93 (10.41) \\
& {IDS-aug-20} & 40.13 (6.12) & 47.96 (7.93) & 46.52 (6.71) & 43.71 (12.65) & 41.38 (8.77) & 29.06 (7.63) & 30.88 (11.74) & 42.63 (13.63) & *40.28 (3.08) \\
\Xhline{1.2pt}
\end{tabular}
\end{table*}

\subsubsection{Comparison with the DataAug Method}\label{sec:exp:DataAug}
The intended usage scenario of DataAug partially overlaps with ours in that a single fully annotated volume (the ``atlas'') for multiclass classification is provided.
However, in order to learn the spatial and appearance transformations for synthesizing extra training data, a pool of unlabeled data are  additionally needed, making DataAug a semi-supervised method.
This difference makes it infeasible to compare DataAug with our proposed MRE-Net in exactly the same setting (in contrast to the comparison with the sSE method);
instead, we formulate a comparison setting as described below.

Noticing that there are 20 scans in the BTCV dataset that are not used in the low-shot experiments and thus can be used as unlabeled data, we compare with DataAug on the BTCV dataset.
In addition, it can be prohibitively time-consuming to train and test DataAug following the one-shot setting described in Section \ref{sec:exp:setting} (\emph{i.e.}, repeating the one-shot experiment by using each of the seven scans selected for low-shot experiments as the atlas in turn), as it necessitates training of three DNN models (spatial transformation, appearance transformation, and segmentation) for just one of these experiments.
Instead, we pick the scan that is closest to the anatomical average of the seven scans as the atlas---as done in \cite{zhao2019dataaug}---and report the performance evaluated on the remaining six scans.
To evaluate the varying benefits of using different amounts of unlabeled data, we experiment with 5, 10, and 20 unlabeled scans for DataAug.
Further, we experiment with two strategies proposed in \cite{zhao2019dataaug} for generating training data with the learned transformation models:
(i) data augmentation using single-atlas segmentation (SAS-aug), where the segmentation mask of the atlas is warped to the unlabeled data, which are then used as additional training data for the segmenter; and
(ii) data augmentation using independent sampling (IDS-aug), where the spatial and appearance transformations are independently sampled and combined to synthesize training data.
The generated training data are used to train the U-Net~\cite{luna20183d} along with online data augmentation.

The experimental results are presented in Table \ref{tab:DataAug}, from which we summarized three findings.
First, increasing the number of unlabeled scans from 5 to 20 consistently improved the performances for both variants of DataAug.
Second, using the U-Net trained with the only annotated sample as baseline, extra training data produced by DataAug helped the segmentation of certain larger organs (spleen and liver) but hurt that of others, ultimately leading to slightly worse cross-organ mean results.
Third, our MRE-Net yielded the best performance for all organs and both metrics. 
We conjecture that the conflicting performance of DataAug was caused by the unsatisfactory registration of the abdominal CT data, as it was originally proposed for registering and synthesizing brain MRI data \cite{zhao2019dataaug}, which intrinsically present subtler individual variations.
Consequently, the unsatisfactory registration resulted in synthesized data/warped masks of low quality, which did not help or even interfered the training of the segmentation networks.
A side evidence is that for larger organs that were registered more reasonably, we did observe apparent performance improvement over the U-Net baseline.
We thus believe that if a spatial transformation model better suited to the abdominal organs could be integrated into the DataAug pipeline, better performance can be expected.


\section{Discussion and Conclusion}
In this work, we proposed a new DNN-based framework, the MRE-Net, for generalized low-shot medical image segmentation in the scenario of extremely scarce training samples and annotations based on DML.
The framework addressed difficulties in this challenging task (most notably data scarcity and overfitting, followed by class imbalance and great diversity) by fully utilizing inter-subject similarities and intraclass variations via multimodal representation learning, as well as incorporating other targeted solutions including structural similarity, online hard example mining, and attentional multi-scale feature embedding.
Experimental results on the MRBrainS18 and BTCV datasets showed that our framework was superior to competing methods including registration in both one-shot and few-shot settings.
As far as the authors are aware of, this work is the first attempt at addressing the extreme scarcity of both annotation and data for DNN-based segmentation, which is expected to exert a significant impact in applicable clinical and research scenarios.

\subsubsection{Clinical Impact}
The scenario considered in this work has great clinical value for rare diseases.
There are about 7,000 known rare diseases~\cite{jia2018rdad} which collectively affect about 400 million people in the world~\cite{molster2018evolution}.
It is generally difficult to obtain data for these conditions which are usually overlooked by the medical imaging community.
For example, Li \emph{et al.} considered as few as one to several samples per class for rare disease diagnosis~\cite{li2020difficulty}.
Although in this work common brain MRI and abdominal CT data were used for experiments, for a valid evaluation, we deliberately used extremely limited data to simulate the real situation---just like Li \emph{et al.} did~\cite{li2020difficulty}---and achieved promising results in our experiments.
Hence, we expect the proposed MRE-Net to work effectively in the target scenario of rare diseases.
In addition, in case of emergency, the MRE-Net can be employed as a rapid yet effective response prior to the accumulation of sufficient data for fully or semi-supervised approaches.

\subsubsection{Performance and Comparison}
ANTs~\cite{avants2011reproducible} and U-Net~\cite{ronneberger2015u,luna20183d} 
are generally considered as benchmarks of classical intensity-based registration and generic DNN-based medical image segmentation, respectively.
In our experiments on MRBrainS18, ANTs achieved decent one-shot results, and the U-Net~\cite{luna20183d} achieved competitive six-shot results.
However, they are known to be either time-consuming or data-hungry.
In this work, we aimed at solving these two issues simultaneously.
On one hand, the proposed MRE-Net outperformed both ANTs and the U-Net (including U-Net trained with times more data) in the low-shot setting.
On the other hand, MRE-Net was orders of magnitudes faster than ANTs (a few seconds versus 28 minutes).
Moreover, on the BTCV dataset,
neither ANTs nor the U-Net managed to produce meaningful results (Table~\ref{tab:BTCV}).
In contrast, our MRE-Net produced quite reasonable results even with only one training sample.
We attribute the superior performance of the MRE-Net to the employment of the DML for low-shot learning as well as the proposal of several targeted solutions considering characteristics of medical images (Table~\ref{tab:ablation}).
It is noticed that the MRE-Net requires more GPU resources for training than the U-Net~\cite{luna20183d} (last row of Table~\ref{tab:implementation}), mainly because the DML necessitates dense distance computations for exhaustive pairwise combinations of the pixel-wise embeddings and category prototypes.
However, considering the substantial performance gains (\emph{e.g.}, up to 124\% increase in mean Dice and up to 80\% reduction in mean HD95 in one-shot settings, and the superior performance of the MRE-Net to the U-Net
in low-shot settings), we believe the trade-off between computational resources and performance is practically acceptable in the realm of low-shot medical image segmentation.

The scenario considered in this work differed from the conventional settings in the literature, making direct comparison with existing methods
difficult.
We focused on the scenario in which both the annotation and data are scarce,
and deliberately used small amounts of data for experiments.
We are aware that existing approaches to low-shot medical image segmentation require auxiliary training data;
however, we think it would still be informative to compare our proposed MRE-Net with some SOTA approaches in controlled settings.
Specifically, we compared with the sSE method~\cite{roy2020squeeze} in the same one-shot setting as ours, and the DataAug method~\cite{zhao2019dataaug} trained in the semi-supervised one-shot setting.
The comparison results emphasized the need for a tailored solution to the specific problem we consider, as both approaches---simply adapting sSE, and applying DataAug to abdominal CT given small to modest amounts of unlabeled data---yield compromised results.
In contrast, our MRE-Net was specifically designed for the extremely low-data regime, and a prominent advantage over existing approaches is the avoidance of large amounts of data.
Another advantage worth mentioning is that the MRE-Net naturally unifies both one- and few-shot learning with a single framework;
in contrast, although extension to few-shot setting is possible for the two compared methods~\cite{zhao2019dataaug,roy2020squeeze}, the optimal way of extension was neither presented nor obvious.

Compared to the non-low-shot performance reported in two SOTA related works \cite{luna20183d,gibson2018automatic},
apparent gaps were observed for the low-shot performance of the MRE-Net on the BTCV dataset (although MRE-Net-1 achieved comparable or even superior results to those of \cite{zhou2016three} for some organs).
Such differences were expected, considering that 80 and 230 times data were used for training in \cite{gibson2018automatic} and \cite{zhou2016three}, respectively.
Note that considering the extremely limited training data in low-shot settings, we chose a relatively lightweight model as compared to the DenseVNet~\cite{gibson2018automatic}, to avoid potential overfitting.
Similarly, wide performance gaps between low-shot and non-low-shot results were also reported in \cite{roy2020squeeze}.
On one hand,
if access to many fully annotated data is provided, it is always recommended to design more powerful models of larger capacity to make full use of the data and annotations.
On the other hand, our work was focused on the data scarcity problem---instead of trying to beat non-low-shot models trained with abundant data and annotations, and our experimental results demonstrated effectiveness of the proposed MRE-Net in producing reasonable segmentation in the target scenario.

\begin{figure}[!t]
\centering
\includegraphics[width=.98\columnwidth,height=5mm,trim=0 20 0 20,clip]{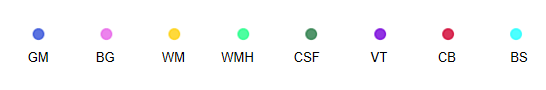}\\
\subfloat[]{\includegraphics[width=.49\columnwidth,trim=71 48 59 51,clip]{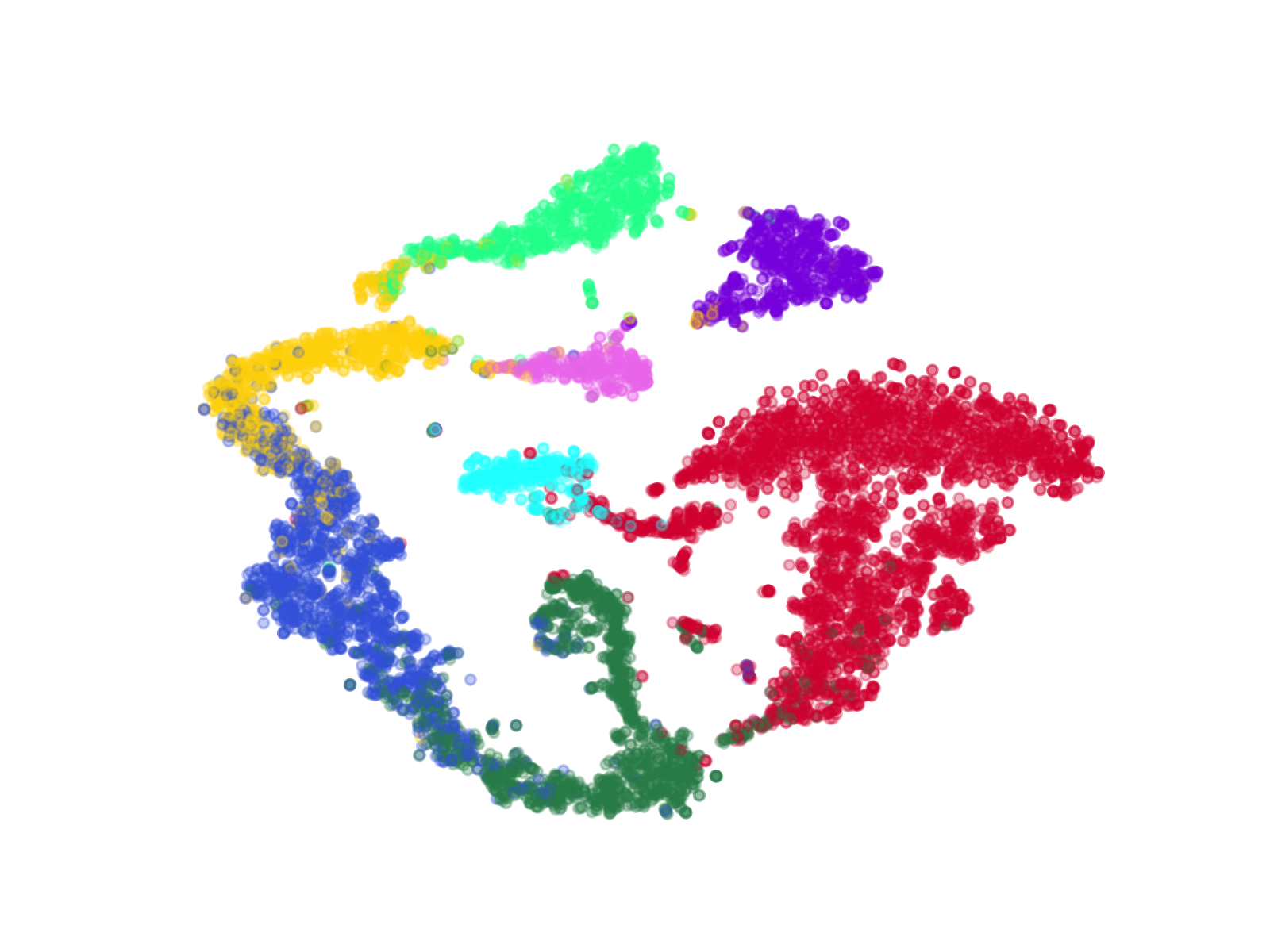}%
\label{fig_first_case}}
\hfil
\subfloat[]{\includegraphics[width=.49\columnwidth,trim=71 47 60 51,clip]{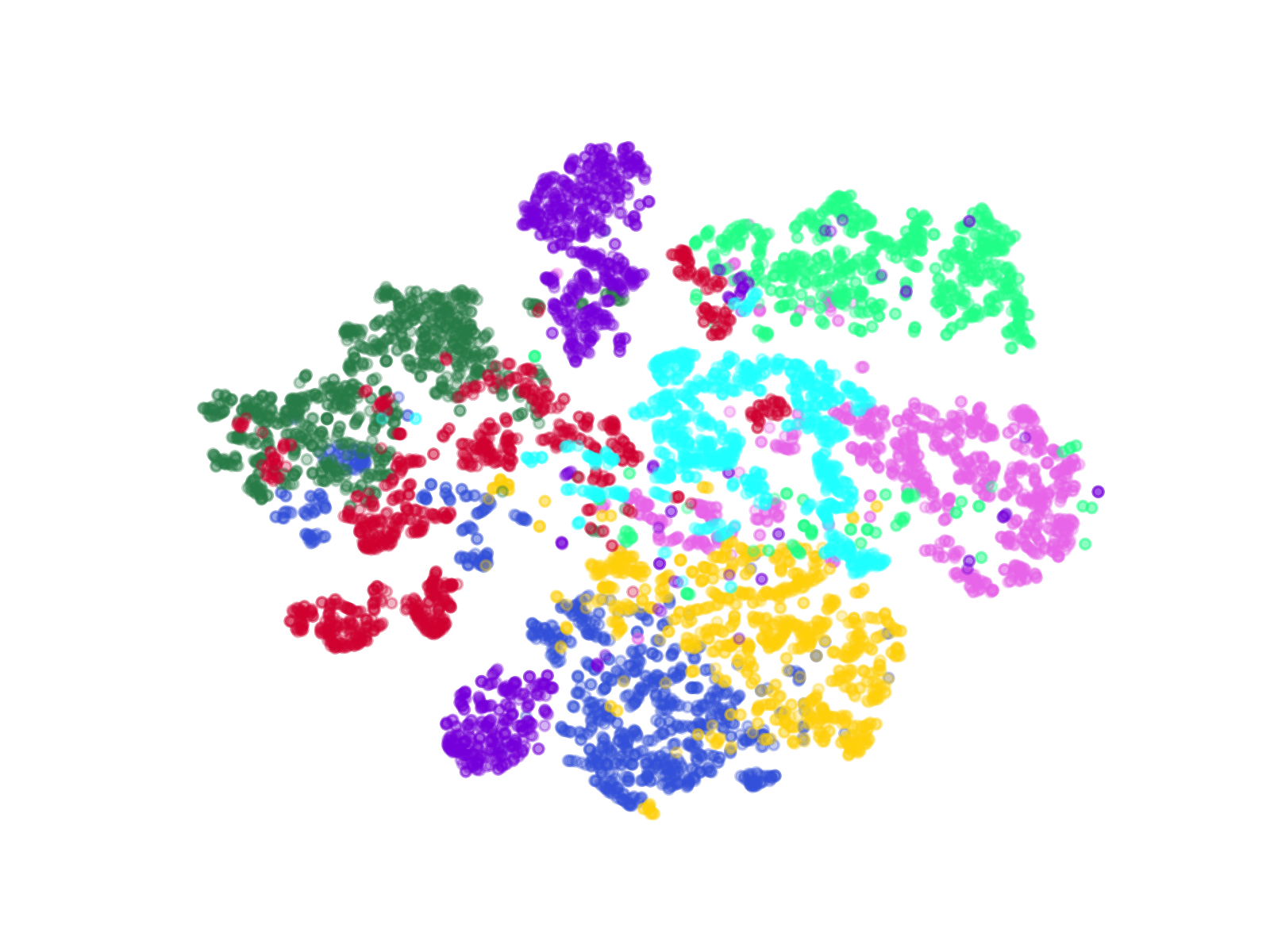}%
\label{fig_second_case}}
\caption{Test sample visualization (t-SNE) of the feature spaces learned by (a) the proposed MRE-Net and (b) the 3D U-Net~\cite{luna20183d} (with online data augmentation) on the MRBrainS18 dataset in one-shot setting.
Different colors represent different tissues (best viewed digitally).}
\label{fig:tsne}
\end{figure}

\subsubsection{Why Does Our MRE-Net Work?}
To effectively utilize the inter-subject similarity and model the intraclass variations from the extremely limited data, our proposed MRE-Net employed the DML to learn a multimodal representation for each category and address the key issue of overfitting.
The inter-subject similarity was utilized by the MRE-Net in the sense that, the DML essentially determined the semantic label of a pixel based on its similarity (distance) to the category prototypes, which were learned from the training data in the embedding space (see Fig.~\ref{fig:concept} and Section~\ref{sec:method:overview}).
This non-parametric approach basically helped prevent overfitting~\cite{snell2017prototypical}.
The critical role of the DML was demonstrated by an ablation experiment, in which the performance went straight down after the DML was removed (row (f) of Table~\ref{tab:ablation}).
For an intuitive interpretation, we employ the t-SNE~\cite{maaten2008visualizing} to visualize the feature embeddings learned by the MRE-Net, in comparison to those by the 3D U-Net~\cite{luna20183d}.
In Fig.~\ref{fig:tsne}(a), data points of different categories formed clear and compact clusters for the MRE-Net, whereas those in Fig.~\ref{fig:tsne}(b) were more difficult to separate for the 3D U-Net.
This visualization demonstrated that the DML helped our MRE-Net learn a better feature space where the embeddings of different categories were more separable (recall that the goal of DML is to make the projected samples of the same class closer to each other than those of different classes in the embedding space).
Meanwhile, our MRE-Net captured the intraclass variations with a mixture of multiple modes for each category, such that the diversity contained in the limited training data was fully exploited.
As empirically demonstrated in Table~\ref{tab:num_mode}, the multimodal framework (\emph{i.e.}, $M=2$ and $3$) gained considerable advantages in performance over the unimodal counterpart ($M=1$), in both one- and three-shot settings.

The proposed framework implemented the concept of multimodal representation learning using the cosine normalization technique~\cite{luo2018cosine}, by embedding the representations as weights of the fc layer instead of allocating specialized memory for them, and computing the cosine similarities via forward prorogation~\cite{qi2018low}.
Correspondingly, we named it the Multimodal Representation Embedding Network (MRE-Net).
In this way, the network design was simplified, and by utilizing the heavily optimized built-in implementation of the fc layer in commonly used deep learning platforms, the DML can become more efficient
(Table~\ref{tab:ablation_distance}).
In addition, the MRE design allowed a unified structure to be applied in both one- and few-shot settings;
that is, the intraclass variations, no matter of a single or few training samples, can be modeled the same way by the MRE.
Along with the use of multimodal representations is how to set the mixing coefficients for the modes.
Instead of the one-hot~\cite{karlinsky2019repmet} or all-equal coefficients, we innovatively proposed an input-dependent mode-weighting strategy based on the self-attention mechanism~\cite{hu2018squeeze}.
As experimentally validated, this strategy improved performance upon the one-hot and all-equal coefficients with adaptable ones (Table~\ref{tab:mixing_coefficients}).


\subsubsection{Application to Ultrasound Images}
We noticed that our experiments so far were all conducted with volumetric CT and MRI images.
In practice, ultrasound imaging also plays an important role in clinics and the segmentation of ultrasound images remains challenging.
For example, Huang \emph{et al.} proposed a novel method based on semantic classification of superpixels (SCS) for tumor segmentation in breast ultrasound (BUS) images, which is difficult due to the poor image quality~\cite{huang2020segmentation}.
To gain an understanding of how our MRE-Net would perform on such challenging tasks of vastly different characteristics (2D segmentation of low-quality images), we applied it to the BUS dataset collected in~\cite{huang2020segmentation}.
This dataset included 220 images for training and 100 for testing, with a relatively small, normalized size of 128$\times$128 pixels.
For few-shot learning, we used 1\%, 5\%, and 10\% of all the training images to train a 2D version of our MRE-Net (replacing 3D operations in the backbone in Fig.~\ref{fig:struct}(a) with 2D counterparts), and compared its performance with that of the FCN~\cite{long2015fully}.
The FCN was trained with the task-optimized hyperparameters described in \cite{huang2020segmentation}.
For evaluation, we adopted the same metrics as in~\cite{huang2020segmentation}, including: averaged radial error (ARE), true positive (TP), false positive (FP), and F1-score.
ARE evaluates object boundary approximation error, whereas TP and FP are region-based and F1-score is a balanced metric between TP and FP (for more details please refer to \cite{huang2020segmentation}).

\begin{table}[!t]
\caption{Few-shot segmentation results on the BUS dataset~\cite{huang2020segmentation} ($\uparrow$: higher is better; $\downarrow$: lower is better).
SCS rept. and expt. represent the numbers originally reported in \cite{huang2020segmentation} and obtained in our experiment, respectively.
Values are mean (std.).
Bold faces denote the overall best numbers for different metrics, and asterisks (*) indicate statistically significant differences from the results of our MRE-Net for each ratio setting.}\label{tab:BUS}
\centering
\setlength{\tabcolsep}{.65mm}
\begin{adjustbox}{width=.99\columnwidth}
\begin{tabular}{cccccc}
\Xhline{1.2pt}
No. images (ratio) & F1-score$\uparrow$ (\%) & ARE$\downarrow$ (\%) & TP$\uparrow$ (\%) & FP$\downarrow$ (\%) \\ \Xhline{0.65pt}
\multicolumn{1}{l}{2 (1\%)} \\
\cline{1-1}
\multicolumn{1}{r}{MRE-Net} & 89.69 (3.55) & 15.95 (14.32) & 87.52 (5.58) & \textbf{7.81} (9.49) \\
\multicolumn{1}{r}{FCN~\cite{long2015fully,huang2020segmentation}} & *79.89 (4.53) & *21.60 (11.45) & *73.07 (6.34) & *9.81 (8.13) \\
\hline
\multicolumn{1}{l}{11 (5\%)} \\
\cline{1-1}
\multicolumn{1}{r}{MRE-Net} & 90.97 (5.45) & 13.87 (15.18) & 93.45 (4.73) & 12.73 (14.93) \\
\multicolumn{1}{r}{FCN~\cite{long2015fully,huang2020segmentation}} & *87.69 (4.82) & *19.59 (15.16) & *92.36 (4.51) & *19.00 (14.58) \\
\hline
\multicolumn{1}{l}{22 (10\%)} \\
\cline{1-1}
\multicolumn{1}{r}{MRE-Net} & \textbf{91.88} (3.97) & 10.49 (9.36) & \textbf{94.17} (4.69) & 11.13 (10.43) \\
\multicolumn{1}{r}{FCN~\cite{long2015fully,huang2020segmentation}} & *88.24 (3.70) & *18.24 (10.70) & *90.97 (4.41) & *15.49 (9.75) \\
\hline
\multicolumn{1}{l}{220 (100\%)} \\
\cline{1-1}
\multicolumn{1}{r}{FCN~\cite{long2015fully,huang2020segmentation}} & 90.49 (2.80) & 12.78 (8.64) & 91.93 (4.75) & 11.36 (7.71) \\
\multicolumn{1}{r}{SCS rept.~\cite{huang2020segmentation}} & 89.97 (4.05) & 9.95 (4.42) & 91.41 (5.04) & 12.22 (9.42) \\
\multicolumn{1}{r}{SCS expt.~\cite{huang2020segmentation}} & 91.17 (2.88) & \textbf{9.81} (6.28) & 93.75 (4.68) & 12.06 (8.10) \\
\Xhline{1.2pt}
\end{tabular}
\end{adjustbox}
\end{table}

The results are shown in Table~\ref{tab:BUS}.
As we can see, our MRE-Net yielded significantly better results than those of the FCN for all metrics in all different ratio settings;
meanwhile, our few-shot results appeared to be quite practical, and generally improved as the amount of training data increased.
In addition, using only 10\% of the training images, our MRE-Net achieved superior or comparable performances with the FCN and SCS approaches trained with all available training images.
These results demonstrated the wide applicability of our MRE-Net to common clinic image data for generalized low-shot segmentation.

\subsubsection{Future Work}
We identified directions for future work based on current limitations.
First, as mentioned in Section~\ref{sec:exp:MRBrainS18}, the WMH is very challenging to segment for its irregular shape, size, or location.
Specifically, we examined the ablation study results of WMH before and after adding the spatial information (rows (b) and (c) in Table~\ref{tab:ablation}).
Contrary to what was expected, WMH also benefited from incorporating the coordinates (Dice $+$4.47\% and HD95 $-$4.02 mm),
suggesting weak spatial preference.
In future work we would like to design more effective strategies exclusively for improving segmentation of irregular abnormalities like the WMH.
Second, we did not explore the impact of pretraining followed by transfer learning in this work.
According to the literature, such a procedure often boosts performance of DNNs in case of limited training data~\cite{tan2018survey}.
Conforming to the specific scenario considered (\emph{i.e.}, extremely scarce training data for the target task), we plan to pretrain the MRE-Net on data of a different task and transfer the learned knowledge onto the target task.
Lastly, the proposed MRE-Net consumes considerable computational resources.
Optimization of the resource consumption is worth future exploration.


%




\ifCLASSOPTIONcaptionsoff
  \newpage
\fi

\IEEEtriggeratref{38}


\bibliographystyle{IEEEtran}
\bibliography{ref}
%

%








\end{document}